%% file: acl_latex.tex
\pdfoutput=1

\documentclass[11pt]{article}

\usepackage[]{acl}

\usepackage{times}
\usepackage{latexsym}
\usepackage{soul}

\usepackage[T1]{fontenc}

\usepackage[utf8]{inputenc}

\usepackage{microtype}

\usepackage{graphicx}
\usepackage{stfloats}
\usepackage{subfigure}
\usepackage{arydshln}
\usepackage[linesnumbered, ruled,vlined]{algorithm2e}
\usepackage{algorithmic}
\usepackage{color}
\usepackage{framed}
\usepackage{xcolor}
\usepackage{bbm}
\usepackage{amsmath}
\usepackage{booktabs}
\usepackage{pifont}

\newcommand{\hs}[1]{\textcolor{blue}{Huan: #1}}
\newcommand{\cl}{\textcolor{brown}}
\newcommand{\wl}{\textcolor{teal}}
\newcommand{\ysu}[1]{\textcolor{cyan}{Yu: #1}}
\newcommand{\zts}[1]{\textcolor{red}{Tianshu: #1}}

\newcommand{\nop}[1]{}

\title{Federated Learning for Semantic Parsing:\\ Task Formulation, Evaluation Setup, New Algorithms}

\author{
    Tianshu Zhang\textsuperscript{1}\thanks{\ \ Work started during the internship at IBM T. J. Watson Research Center and continued at OSU.}, Changchang Liu\textsuperscript{2}, Wei-Han Lee\textsuperscript{2}, Yu Su\textsuperscript{1}, Huan Sun\textsuperscript{1}  \\ 
    \textsuperscript{1}The Ohio State University\\
    \textsuperscript{2}IBM Research
    \\
	\texttt{\textsuperscript{1}\{zhang.11535, su.809, sun.397\}@osu.edu} \\
	\texttt{\textsuperscript{2}\{changchang.liu33, wei-han.lee1\}@ibm.com}
}

\begin{document}
\maketitle

\input{abstract-v2.tex}
\input{intro-v2.tex}
\input{problem_formulation-v2.tex}
\input{benchmark_analysis-v1.tex}

\input{background-v2.tex}

\input{proposed_method-v1.tex}
\input{experiments-v1.tex}

\input{related_work-v1.tex}

\input{conclusion.tex}

\section*{Limitations}
In this work, we address the heterogeneity challenge in the task of FL for semantic parsing, by leveraging the reduction of training loss signal. Our work is motivated from the FL training procedure perspective to adjust the contribution of each client during the global model aggregation stage, but how each client's data contribute to the final global model is still unclear. As the data of different clients contain different information, what kind of information of each client is helpful and can be more directly linked and utilized to facilitate the FL training is worth more efforts in future work.

In addition, our proposed re-weighting mechanism is a universal technique for cross-silo FL. Thus generalizing our proposed re-weighting mechanism to a broader range of tasks beyond semantic parsing, and further studying under what kind of conditions, \texttt{Lorar} can make a huge difference for FL
would be interesting future work to pursue.

\section*{Acknowledgements}

\nop{\hs{@changchang, feel free to add things from IBM side. We should also say sth like the work started during Tianshu's IBM internship.}}
The authors would like to thank colleagues from
the OSU NLP group and all anonymous reviewers for their thoughtful comments.
This research was supported in part by NSF OAC 2112606, NSF IIS 1815674, NSF CAREER 1942980, and Ohio Supercomputer Center \cite{OhioSupercomputerCenter1987}. The work done at IBM research was sponsored by the Combat Capabilities Development Command Army Research Laboratory and was accomplished under Cooperative Agreement Number W911NF-13-2-0045 (ARL Cyber Security CRA). The views and conclusions contained in this document are those of the authors and should not be interpreted as representing the official policies, either expressed or implied, of the Combat Capabilities Development Command Army Research Laboratory or the U.S. Government. The U.S. Government is authorized to reproduce and distribute reprints for Government purposes notwithstanding any copyright notation here on. We thank Chaoyang He for his help during reproducing FedNLP. We thank Wei-Lun (Harry) Chao for valuable discussion.


\bibliography{anthology,custom}

\newpage
\appendix

\section{Appendix}
\label{sec:appendix}

\nop{
\begin{figure}[h]
\centering
\includegraphics[width=0.5\textwidth]{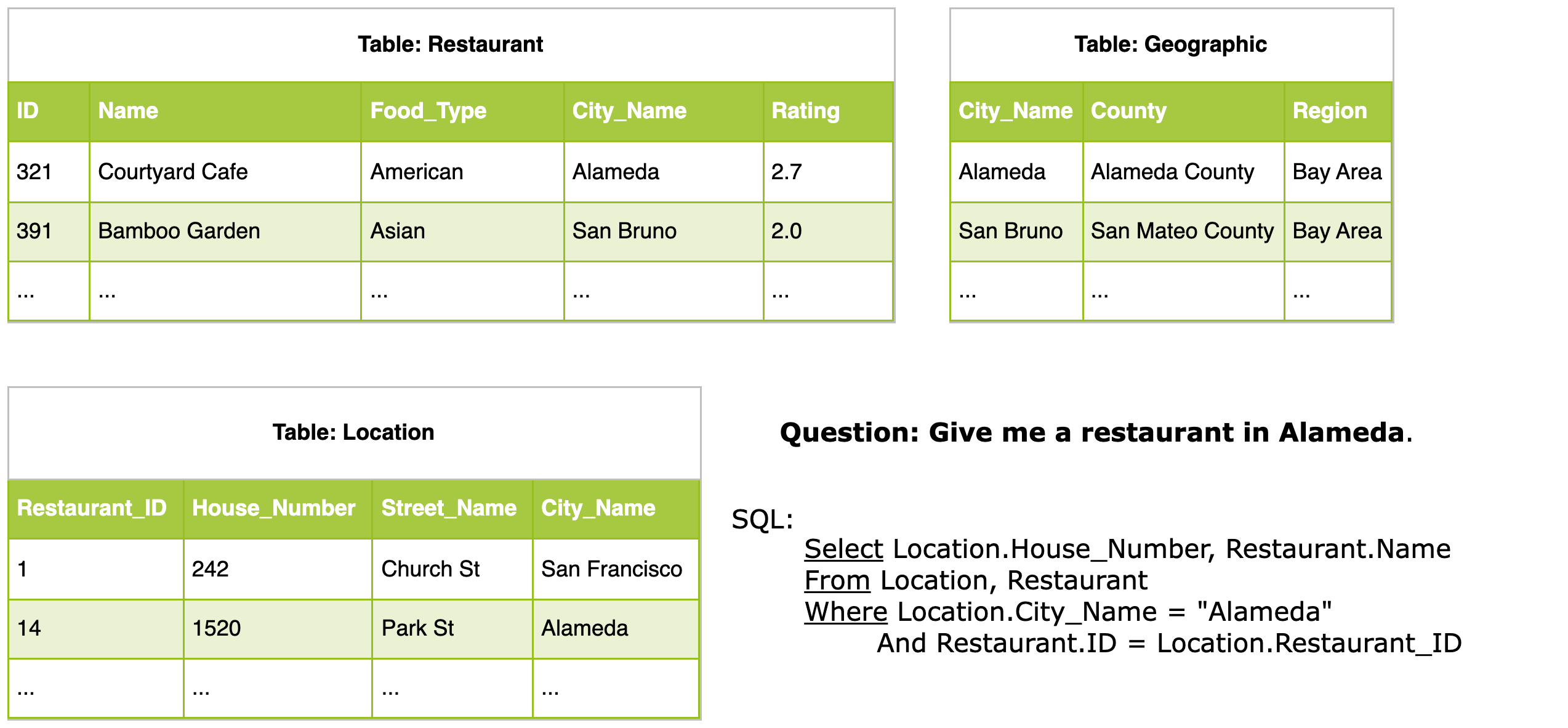}
\caption{An overview of the text-to-SQL task.}
\label{text2sql}
\end{figure}
}

\renewcommand{\thefootnote}{\ding{\numexpr192+\value{footnote}}}
\begin{algorithm}
\caption{}
\label{algorithm}
\KwIn{local datasets $\mathcal{D}_i$, number of communication rounds $T$, number of local epochs $E$, server learning rate $\eta$, client learning rate $\eta_i$}
\KwOut{the final global model $w^T$}
\textbf{Server executes:}\\
\For{$t \in 0,1,2,...,T$}
{   
Sample a set of clients $C_t$\footnotemark[1]{}\\
\For{$i \in C_t$ \textbf{in parallel}}
{
Send the global model $w^t$ to client $i$ \\
$\Delta w_i^t$, $\left|\mathcal{D}_i\right|\Delta \mathcal{L}_i^t$  $\leftarrow$\textbf{LocalTraining$(i, w^t)$}  \\
}
$\Delta w^t = \sum_{i \in C_t} p^t_i \Delta w_i^t $\\
\setlength\FrameSep{0.5ex}
\setlength\OuterFrameSep{0pt}
\definecolor{shadecolor}{RGB}{204,204,255}
\begin{shaded}
For FedOPT/FedAvg/FedProx: $p_i = \left|\mathcal{D}_i\right|/\sum_{i \in C_t} \left|\mathcal{D}_i\right|$ \\
\end{shaded}
\definecolor{shadecolor}{RGB}{204,255,204}
\begin{shaded}
For ours (\texttt{Lorar}): 
$p_i^t = \left|\mathcal{D}_i\right|\Delta \mathcal{L}_i^t / \sum_{i \in C_t} \left|\mathcal{D}_i\right|\Delta \mathcal{L}_i^t$  \\
\end{shaded} 
$w^{t+1} \leftarrow w^t-\eta\Delta w^t$ 
} 
return $w^T$ \\
\textbf{Client executes:} \\
FedAvg/FedOPT: $\mathcal{L}(w;b) = \sum_{(x,y) \in b} f(w;x;y)$ \\
FedProx: $\mathcal{L}(w;b) = \sum_{(x,y) \in b} f(w;x;y) + \frac{\mu}{2}\Vert w-w^t\Vert^2$ \\
\textbf{LocalTraining($i, w^t$)} \\
$w_i^t \leftarrow w^t$ \\
\For{epoch $k = 0,1,2,...,E$}
{
\For{each batch $b = \{x,y\}$ of $\mathcal{D}_i$}
{
$w_i^t \leftarrow w_i^t - \eta_i \nabla \mathcal{L}_i^{t,k}(w_i^t;b)$
}
}
$\Delta w_i^t \leftarrow w^t - w_i^t$ \\
$\Delta \mathcal{L}_i^t \leftarrow \max \mathcal{L}_i^t - \min \mathcal{L}_i^t$ \\
return $\Delta w_i^t, \left|\mathcal{D}_i\right|\Delta \mathcal{L}_i^t$ to the server
\end{algorithm}
\footnotetext{We use all clients in our experiments.}

\nop{\cl{check Line 21 for the three different formats of $L$ and double check with Eqn (6) for consistency}}
\nop{\cl{1. Line 2 t starts from 0 and in Line 17 k starts from 1, let's try to be consistent unless we have any special reasons. 2. For line 7, should we use $p_i^t$ here? 3) For Line 10, let's be consistent about whether to use bold format. Also be consistent with equations in main body. 4) Revise Line 13 and Line 14 to be consistent with Eqn. (1) and (5)}}

\nop{
\begin{table*}[h]
\centering
\resizebox{\linewidth}{!}{
\begin{tabular}{@{}lcccccccccc@{}}
\toprule
\text{} & \text{Advising} & \text{ATIS} & \text{GeoQuery} & \text{Restaurants} & \text{Scholar} & \text{Academic} & \text{IMDB} & \text{Yelp} & \text{MacroAvg} & \text{MicroAvg}\\
\cmidrule(lr){2-9}\cmidrule(lr){10-11}
FedOPT & 77.14 & 51.90 & \textbf{71.68} & \textbf{98.65} & \textbf{63.76} & 
\textbf{26.32} & 23.08 & 12.5 & 53.13 & 65.81\\
FedOPT$_{lorar}$ & \textbf{75.22} & \textbf{54.81} & \textbf{71.68} & \textbf{98.65} & 62.84 & 0 & \textbf{50} & \textbf{29.17} & \textbf{55.30} & \textbf{65.87}\\
\bottomrule
\end{tabular}}
\caption{\label{random epoch}
Random local epochs for FedOPT and FedOPT$_{lorar}$ on the test set. 
}
\end{table*}
}

\subsection{Implementation Details}\label{implementation details}
\nop{{\cl{The structure of this subsection is a bit unclear. Let's reorganize it and break into the following points: 1) Text2SQL model details (T5 and related info in Paragraph 1), 2) Federated algorithm details (Paragraphs 4 and 1), 3) Centralized details (paragraph 3), 4) Finetuning details (paragraph 2), 5) Hyperparameter settings (Paragraphs 5 and 6, part of paragraph 1), 6) Implementation platform}} \zts{I have restructured this section. I didn't change paragraph 5 since those hyperparameters are common for all three settings.}}

\nop{We develop our experiments \hs{you need to be specific: which part is developed based on FedNLP and which part corresponds to UnifiedSKG? see the next paragraph, if it looks good, we could delete this paragraph.} based on FedNLP \cite{lin-etal-2022-fednlp}, FedML \cite{chaoyanghe2020fedml} and UnifiedSKG \cite{UnifiedSKG}.} 

We use T5-base \cite{raffel2020exploring} as the model for text-to-SQL task in all three learning paradigms (finetuning, centralized and FL), as it has been shown as an effective unified model for various semantic parsing tasks in UnifiedSKG \cite{xie-etal-2022-unifiedskg}. For all three FL algorithms, we implement them based on FedNLP \cite{lin-etal-2022-fednlp} and FedML \cite{chaoyanghe2020fedml}. We use Adafactor \cite{pmlr-v80-shazeer18a} as the optimizer for finetuning and centralized paradigms, and as the client optimizer\footnote{Note we use Adafactor as the local optimizer for FedAvg, so the FedAvg in our paper is slightly different from the original proposed FedAvg, which uses stochastic gradient descent(SGD) as the local optimizer.} for FL paradigm, since it has been shown as the best optimizer to optimize for the T5 model. 

For the FL paradigm, we tune hyperparameters for FedOPT, FedAvg and FedProx as follows. For FedOPT, we test all the combinations of the server learning rate from \{0.001, 0.01 0.1, 0.5, 1\} and \{w/ 0.9, w/o\} server momentum. We found 1 as the server learning rate and 0.9 as the server momentum is the best hyperparameter combination. For FedProx, we vary $\mu$ from \{0.0001, 0.001, 0.01, 0.1, 1\} and use the dev set to choose the best model. We finally choose the best hyperparameter 0.0001 in our experiment. For all the federated learning paradigms, we set local training epochs as 6 for two large datasets: ATIS and Advising. We set the local training epoch as 12 for all the other six datasets. We let all the clients participate in each round and we train the entire process for 60 rounds (which lasts around 60 hours).\nop{\hs{maybe minor, but why 62? this number is a bit odd. Why not 60 or 65?}} And we test the global model performance on the merged dev set for every 5 communication rounds to choose the best model. \nop{\hs{Minor question: Does it make sense for each client to use its own dev set to select different global models (i.e., different check points) as its best model? Do we have to use the same global model for each client?}}We use the best global model to evaluate on all eight test sets to get the global model performance on each client.

For the finetuning paradigm, we finetune T5-base on each dataset for a maximum of 200 epochs. We use the dev set of each client to choose the best model and then evaluate the model on each test set.

For the centralized paradigm, we merge all eight training sets and then finetune T5-base for a maximum of 200 epochs on the merged dataset to get one centralized model. We merge all eight dev sets and use the merged dev set to choose the best model. Then we evaluate the centralized model on each test set.

For all finetuning, centralized and federated learning paradigms, we set the input length as 1024 and the output length as 512. We try learning rate in \{1e-5, 1e-4, 1e-3\}. We finally choose 1e-4 for the centralized paradigm, and 1e-4 for Advising, ATIS, Geoquery and Yelp in the finetuning paradigm and FL paradigm. We use 1e-3 for Restaurants, Scholar, Academic and IMDB in the finetuning paradigm and FL paradigm.

\nop{With the hyperparameters we set, for all finetuning, centralized and federated learning paradigms, the model has converged in our experiments \hs{how did you define convergence here?}.\zts{I'll remove this part. It's not fully converge for FedAvg and FedProx.}}

For the computing resources, we use 1 NVIDIA A6000 48GB GPU for finetuning, with batch size 8. We use 2 NVIDIA A6000 48GB GPUs for centralized training, with batch size 8. We use 5 NVIDIA A6000 48GB GPUs for all federated learning experiments. Specifically, one GPU is used as the server and the other four GPUs are used as 8 clients, with each GPU accommodating 2 clients. The batch size for clients GeoQuery, Restaurants, Scholar, Academic, IMDB and Yelp is 4, and for clients Advising and ATIS is 8. 

\begin{figure}[h]
\centering
\includegraphics[width=0.48\textwidth]{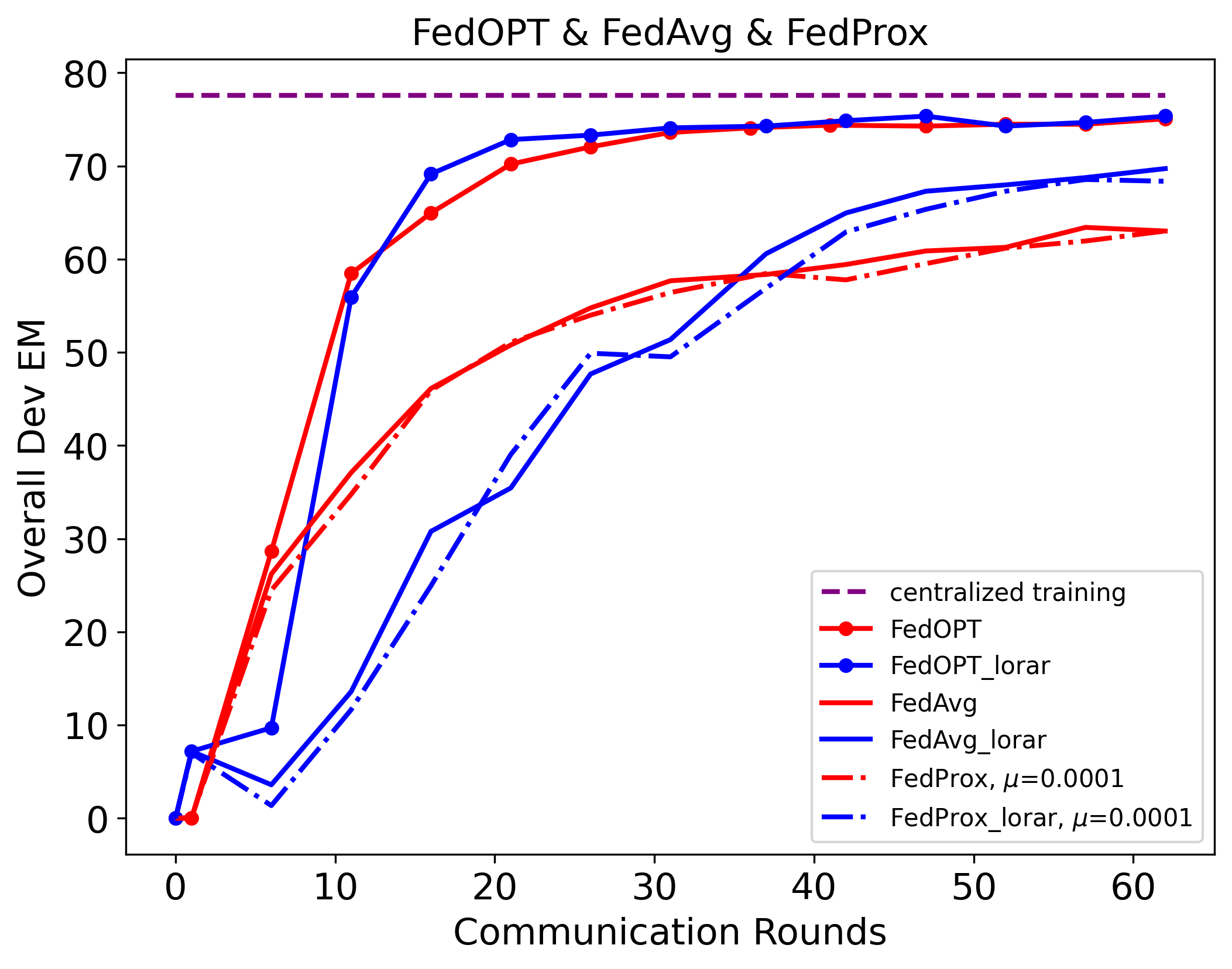}
\caption{Overall dev performance, also equivalent to the MicroAvg on eight clients' dev sets. All the red curves show the original FL algorithms. All the blue curves show the algorithms after applying \texttt{Lorar}. }
\label{overall performance}
\end{figure}

\subsection{Comparison of FL Baselines.} \label{FL baselines comparison}
We treat FedAvg, FedOPT and FedProx as our FL baselines. As Figure \ref{overall performance} shows, among FedAvg, FedOPT and FedProx, FedOPT performs the best, achieving the closest performance to the centralized paradigm and the fastest convergence speed. FedAvg and FedProx have similar performances, and both of them have a large gap with FedOPT\nop{\cl{and both of them performs worse than FedOPT with a significant degradation'}}. This indicates that the server's adaptive optimizer which only exists in FedOPT plays an important role to improve the performance. \nop{And the regularization existing in FedProx which minimizes the distance between the global model and clients' local models does not make too much difference in our case. This may be because the L2 distance is weak to constrain  a model with huge parameters such as T5.}
\subsection{Performance Variation under Varying Communication Rounds.} \label{performance changes}
In Figure \ref{overall performance}, comparing the performance of FL baselines with ours, FedOPT$_{\texttt{lorar}}$ performs slightly better than FedOPT. We hypothesize the small gap between FedOPT and the centralized paradigm limits the room for \texttt{Lorar} to show a large gain over FedOPT. For FedAvg and FedProx, we can see that applying \texttt{Lorar} performs significantly better, which demonstrates the effectiveness of leveraging the loss reduction to adjust the weights.

\nop{
\subsection{Robust to Local Training Epochs}\label{random epoch section}

As FL is sensitive to the local epochs \cite{li2021model}, we also study how our proposed method is robust to the local training epochs. We simulate the extreme setting where during each round, different clients use a random epoch in the local training stage. The local epoch for each client is randomly selected thus different in each round's local training stage. As Table \ref{random epoch} shows, our proposed algorithm FedOPT$_{lorar}$ performs better than FedOPT on both MacroAvg and MicroAvg, which demonstrates that our loss reduction adjusted re-weighting mechanism can be more robust to the local epochs. 
\zts{Implementation Details: For each round, we randomly select epochs from {3,4,5,6} for Advising and ATIS, and from {3,4,5,6,7,8,9,10} for the rest of the six datasets. We use the same selected random epochs for both FedOPT and FedOPT$_{\texttt{Lorar}}$ to run 60 rounds.}
}

\begin{figure}[t]
\centering
\includegraphics[width=0.5\textwidth]{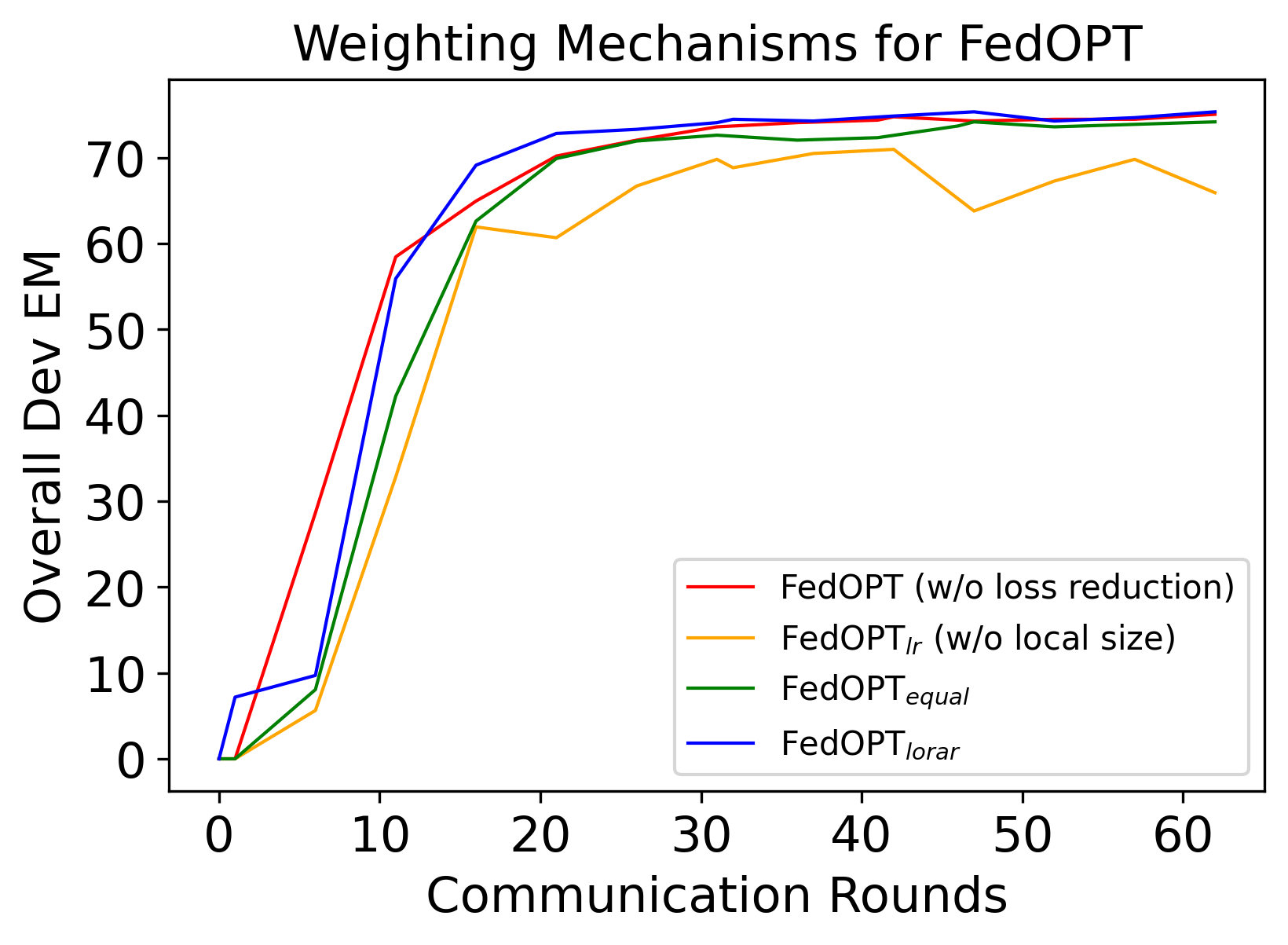}
\caption{Alternative weighting mechanisms for FedOPT on the dev set of our proposed benchmark. Recall that FedOPT uses a client's training set size (w/o loss reduction) as its weight, FedOPT$_{lr}$ refers to only using a client's loss reduction during each round (w/o train set size) as its weight, while FedOPT$_{\texttt{lorar}}$ considers both factors (Eqn. (\ref{simplified_weight})). FedOPT$_{equal}$ means each client gets equal weight.}
\label{fedopt_ablation_study}
\end{figure}

\begin{figure}[h]
\centering
\includegraphics[width=1\textwidth]{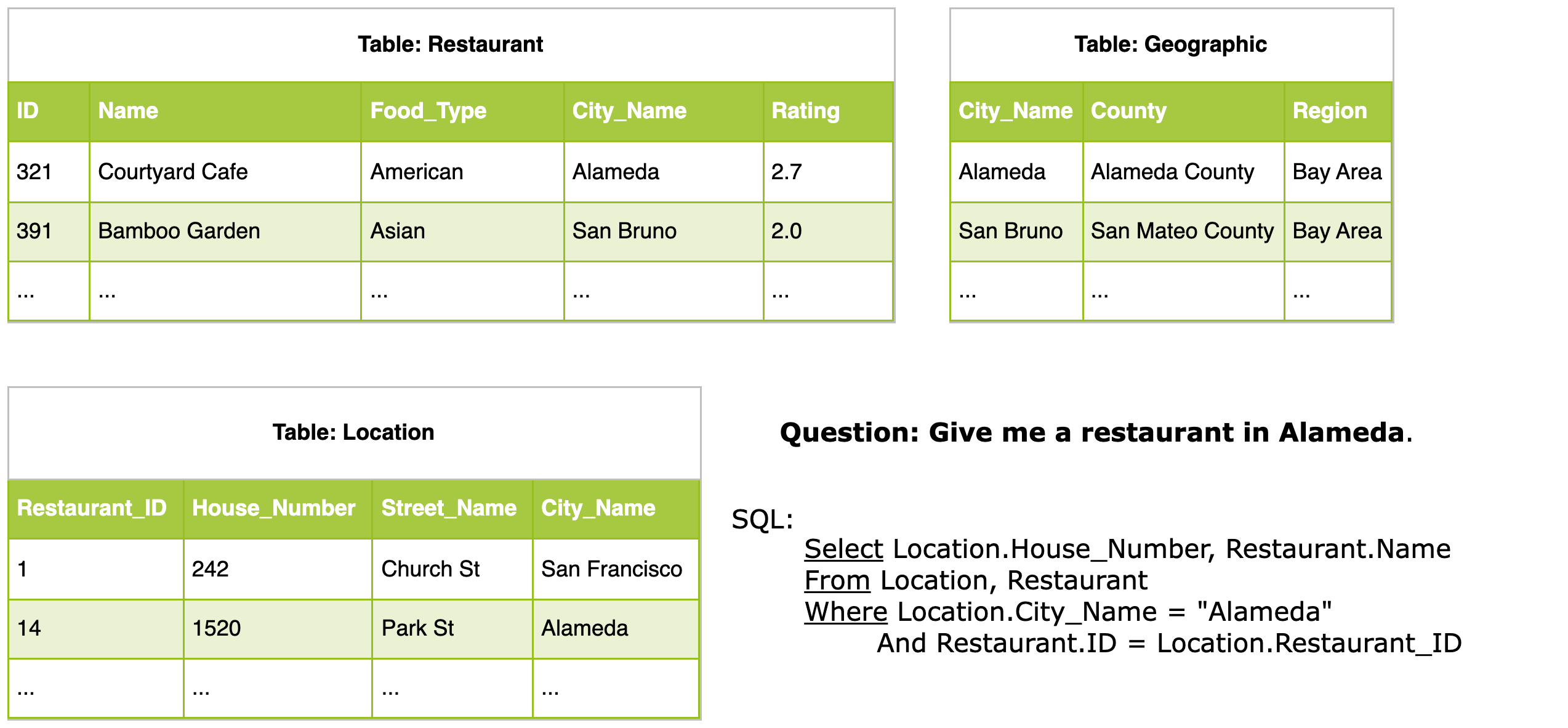}
\caption{An overview of the text-to-SQL task.}
\label{text2sql}
\end{figure}

\end{document}

%% file: abstract-v2.tex
\begin{abstract}
This paper studies a new task of federated learning (FL) for semantic parsing, where multiple clients collaboratively train one global model without sharing their semantic parsing data. By leveraging data from multiple clients, the FL paradigm can be especially beneficial for clients that have little training data to develop a data-hungry neural semantic parser on their own. We propose an evaluation setup to study this task, where we re-purpose widely-used single-domain text-to-SQL datasets as clients to form a realistic heterogeneous FL setting and collaboratively train a global model. \nop{\zts{should we clearly mention "benchmark"? actually it's one important contribution, I feel the "evaluation setup" doesn't reflect the "benchmark" meaning. Also even for MRQA, FedNLP can call it "benchmark"}. }
As standard FL algorithms suffer from  the high client heterogeneity in our realistic setup,
we further propose a novel \underline {\textbf{LO}}ss \underline{\textbf{R}}eduction \underline{\textbf{A}}djusted \underline{\textbf{R}}e-weighting (\texttt{Lorar}) mechanism to mitigate the performance degradation, which adjusts each client's contribution to the global model update based on its training loss reduction during each round. Our intuition is that the larger the loss reduction, the further away the current global model is from the client's local optimum,\nop{\zts{our intuition is more like this, but when we are doing exp, some clients' training losses during the local training stage don't achieve the lowest point (I mean the training loss doesn't achieve the plateau, whether we can still call it "local optimum"?, I think maybe "the best local model" is better}} and the larger weight the client should get.\nop{during each round.}\nop{\zts{By adjusting the weights just based on the data size proportion which weakens the importance of the clients with smaller datasets, \texttt{Lorar} can effectively improve their performance without nearly hurting the performance of the clients with larger datasets.}}\nop{By giving a higher weight to clients that converge more slowly and harder, we implicitly increase the difficulty of the entire training process \hs{this sentence is very important and needs to be more precise} and indirectly enforce the global model toward the direction that takes more care of those clients. \ysu{These two sentences about Lorar are not very precise and convincing. Phrases like ``converge more slowly and harder'', ``implicitly'', ``indirectly enforce'', ``takes more care of'' are vague and weak. Also, this should start with why existing methods are insufficient.}} By applying \texttt{Lorar} to three widely adopted FL algorithms (FedAvg, FedOPT and FedProx), we observe that their performance can be improved substantially on average (4\%-20\% absolute gain under MacroAvg) and that clients with smaller datasets enjoy larger performance gains. {In addition, the global model converges faster for almost all the clients.}\footnote{Our code and data are publicly available at \url{https://github.com/OSU-NLP-Group/FL4SemanticParsing}} 


\end{abstract}

%% file: intro-v2.tex
\section{Introduction}

Semantic parsing aims to translate natural language utterances into formal meaning representations such as SQL queries and API calls and can be applied to build natural language interfaces that enable users to query data and invoke services without programming \cite{berant-etal-2013-semantic, thomason2015learning, su2017building, campagna2017almond}. Neural semantic parsers have achieved remarkable performance in recent years \cite{wang-etal-2020-rat,rubin-berant-2021-smbop, Scholak2021:PICARD}. However, they are data-hungry; bootstrapping a neural semantic parser by annotating data on a large scale can be very challenging for many institutions, as it requires the annotators to have intimate knowledge of formal programs.  One natural thought is to leverage data from different institutions and train a unified model that can be used for all institutions. However, in practice, institutions such as hospitals, banks, and legal firms are prohibited from sharing their data with others, due to privacy concerns. Therefore, for institutions that only have very limited data, it is extremely hard to build their own neural semantic parsers.

Federated learning (FL) \cite{konevcny2016federated, pmlr-v54-mcmahan17a, yang2018applied} has turned out to be a popular training paradigm where multiple clients can collaboratively train a global model without exchanging their own data. In this paper, we study a new task of federated learning for semantic parsing. Through FL on the data scattered on different clients (e.g., institutions), we aim to obtain a global model that works well for all clients, especially those that have insufficient data to build their own neural models.

Towards that end, we propose an evaluation setup by re-purposing eight existing datasets that are widely adopted for text-to-SQL parsing, such as ATIS \cite{data-atis-geography-scholar} and Yelp \cite{data-sql-imdb-yelp}. These datasets demonstrate great heterogeneity, in terms of dataset sizes, language usage, database structures, and SQL complexity, as they were collected from the real life by different researchers, at different times, and for different purposes. Therefore, we use this collection to simulate a realistic scenario where eight clients with very different data participate in the FL paradigm to jointly train a neural semantic parser.

Heterogeneity, where the data distributions and dataset sizes on different clients are different, is recognized as one of the biggest challenges in FL \cite{pmlr-v54-mcmahan17a, reddi2020adaptive, li2020federated, li2021fedbn, shoham2019overcoming, t2020personalized}.  Existing work either uses synthetic data \cite{li2020federated} or splits a classification dataset based on Dirichlet distribution \cite{lin-etal-2022-fednlp} to simulate the non-IID federated learning setting,\nop{ and even FedNLP \cite{lin-etal-2022-fednlp} uses MRQA \cite{} to simulate the} while we propose a more realistic setup to study this setting for semantic parsing. Pre-trained language models such as T5 \cite{raffel2020exploring} have been shown as a powerful unified model for various semantic parsing tasks \cite{xie-etal-2022-unifiedskg, rajkumar2022evaluating}, which can be leveraged to save us the efforts for client-specific model designs. Specifically, we adopt T5-base as our backbone semantic parser in the FL paradigm, and conduct extensive experiments and analysis using three widely-adopted FL algorithms: FedAvg \cite{pmlr-v54-mcmahan17a}, FedOPT \cite{reddi2020adaptive} and FedProx \cite{li2020federated}.

\begin{figure}[t!]
\centering
\includegraphics[width=0.49\textwidth]{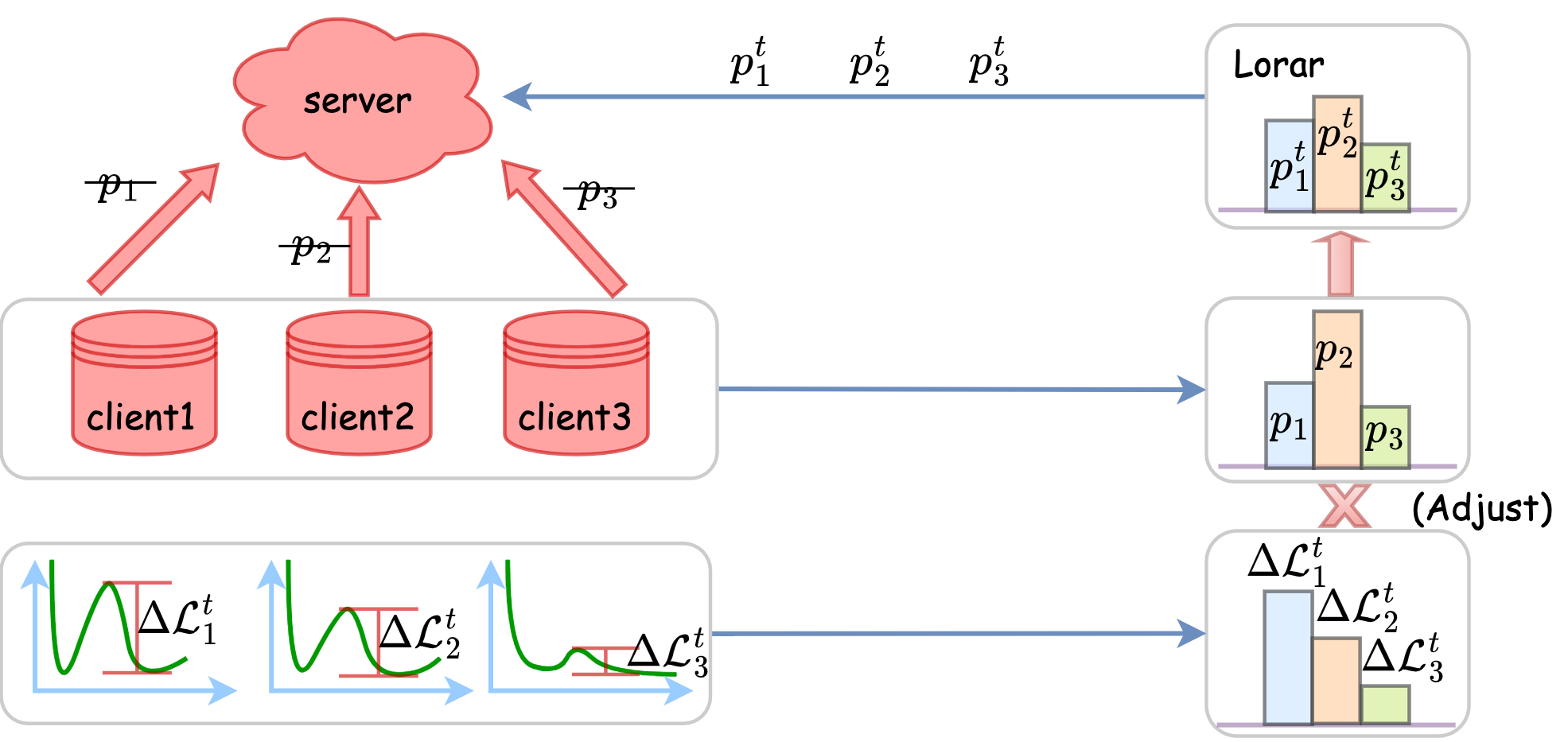}
\caption{Our proposed re-weighting mechanism \texttt{Lorar} for the global model update in each round. The weight for each client (i.e., its contribution to the global model update) will be adjusted based on its loss reduction in each round. $\Delta \mathcal{L}^t_i$ means the training loss reduction of the \textit{i-th} client in the \textit{t-th} round. \nop{{\cl{This figure may needs further modification. Firstly, we need to describe the original weight pi? Secondly, the current figure easily make reviewer misunderstood that the weights are calculated collectively from all clients, instead of individually calcaulated by each client.}}}\nop{\hs{is the `+' symbol on the right correct?} {\wl{To make this figure clearer and consistent with Eq. 11, let's plan to change $p1, p2, p3, p1', p2', p3'$  to $p_1^t, p_2^t, p_3^t, p_1^{t+1}, p_2^{t+1}, p_3^{t+1}$. Please also make necessary changes to other notations as well to be consistent}}}\nop{\cl{Let's add two more figures of the new training losses after applying the new method. \ysu{this comes too late. It needs to be on parge 1 and referenced multiple times in the intro}} \hs{I think it's ok to put it on page 2. This isn't the contribution that is worth the most highlight.}}}
\label{motivation_example}
\end{figure}

As standard FL algorithms suffer from the high client heterogeneity in our realistic setup, we further propose a novel re-weighting mechanism for combining the gradient updates from each client during the global model update. The high-level idea is shown in Figure~\ref{motivation_example}. Our intuition is that, for each client, the reduction of training loss during each round can signalize how far the current global model is away from the local optimum. By giving larger weights to those clients that have larger training loss reduction, the global model update can accommodate those clients better, thus mitigating potential performance degradation caused by high heterogeneity. We formulate this intuition as a re-weighting factor to adjust how much each client should contribute to the global model update during each round. Our proposed mechanism can be applied to all the three FL algorithms and experiments show that it can substantially improve both their parsing performance and their convergence speed, despite being very simple.

In summary, our main contributions are:
\begin{itemize}
\item [$\bullet$]To the best of our knowledge, we are the first to study federated learning for semantic parsing, a promising paradigm for multiple institutions to collaboratively build natural language interfaces without data sharing, which is especially beneficial for institutions with little training data. 
\item [$\bullet$]We propose an evaluation setup to simulate a realistic heterogeneous FL setting where different participating institutions have very different data. We re-purpose eight single-domain text-to-SQL datasets as eight clients, which demonstrate high heterogeneity in terms of dataset sizes, language usage, database structures, and SQL complexity. 
\item [$\bullet$]We propose a novel re-weighting mechanism, which uses the training loss reduction of each client to adjust its contribution to the global model update during each round. Experiments show that our re-weighting mechanism can substantially improve the model performance of existing FL algorithms on average, and clients with smaller training data observe larger performance gains. We discuss the limitations of our work and encourage future work to further study this task.
\end{itemize}

%% file: problem_formulation-v2.tex
\section{Motivation and Task Formulation}
\nop{\hs{follow the logic in intro to briefly talk about why we want to study this task.}}

\nop{Semantic parsing is a task that tries to translate natural language utterances into 
programs.\hs{I think you could just say semantic parsing and introduce the definition about text-to-SQL in the next section after saying you use text-to-SQL datasets for evaluation. Again, check the logic in Intro.}} 

Semantic parsing aims to translate natural language utterances into formal meaning representations and has numerous applications in building natural language interfaces that enable users to query data and invoke services without programming. As many institutions often lack data to develop neural semantic parsers by themselves, we propose a federated learning paradigm, where clients (i.e., ``institutions'') collaboratively train a global semantic parsing model without sharing their data. 


\nop{\hs{the connection between these two paragraphs is not fluent. you should also highlight you are the first one to study this task.}
Federated Learning is a training paradigm, which multiple clients collaboratively train a model without sharing their data. }
There are two realistic settings of FL: cross-silo setting and cross-device setting \cite{kairouz2021advances, lin-etal-2022-fednlp}.\nop{\hs{give citations!!}} For the cross-silo setting, clients are large institutions, such as hospitals and companies, and the number of clients is limited in this setting. In general, they have large computational resources and storage to train and store a large model, and large communication costs between the server and clients are tolerated.\nop{And it can also tolerate huge communication costs between the server and clients.} For the cross-device setting, clients are small devices such as mobile phones and Raspberry Pis, thus there may exist a huge number of clients. They have limited computational resources and storage and only small communication costs between the server and clients are affordable. Here our FL for semantic parsing can be regarded as a cross-silo setting, where each client is a relatively large institution that hopes to build a natural language interface based on its user utterances and underlying data. Studying FL for semantic parsing under a cross-device setting could be interesting future work.


%% file: benchmark_analysis-v1.tex
\section{Evaluation Setup}
\label{benchmark}
\begin{table*}[t]
\centering
\resizebox{\linewidth}{!}{
\begin{tabular}{@{}l|c|ccc|ccccccc@{}}
\toprule
\text{} & \text{} & \text{} & \text{} & \text{} & \text{SQL} & \text{Questions} & \multicolumn{2}{c}{Unique tables} & \multicolumn{2}{c}{\texttt{SELECT}s}\\
\text{} & \text{Domain} & \text{Train} & \text{Dev} & \text{Test} & \text{Pattern} & \text{/ unique query} & \multicolumn{2}{c}{/ query} & \multicolumn{2}{c}{/ query} \\
\text{} & \text{} & \text{} & \text{} & \text{} & \text{count} & \text{count} & \text{$\mu$} & \text{Max} & \text{$\mu$} & \text{Max} \\
\midrule
Advising & Course Infomation & 2629 & 229 & 573 & 174 & 21.7 & 3.0 & 9 & 1.23 & 6 \\
ATIS & Flight Booking & 4347 & 486 & 447 & 751 & 5.6 & 3.8 & 12 & 1.79 & 8 \\
GeoQuery & US Geography & 549 & 49 & 279 & 98 & 3.6 & 1.1 & 4 & 1.77 & 8 \\
Restaurants & Restaurants/Food& 228 & 76 & 74 & 17 & 16.4 & 2.3 & 4 & 1.17 & 2 \\
Scholar & Academic Publication& 499 & 100 & 218 & 146 & 4.2 & 3.2 & 6 & 1.02 & 2 \\
Academic & Microsoft Academic& 120 & 38 & 38 & 92 & 1.1 & 3 & 6 & 1.04 & 3 \\
IMDB & Internet Movie & 78 & 26 & 26 & 52 & 1.5 & 1.9 & 5 & 1.01 & 2  \\
Yelp & Yelp Website& 78 & 26 & 24 & 89 & 1.2 & 2 & 4 & 1 & 1 \\
\bottomrule
\end{tabular}}
\caption{\label{data statistics}
Statistics for the heterogeneous text-to-SQL datasets. "$\mu$": the average number under the measure. "Max": the max number under the measure.}
\end{table*}

As we are the first to study cross-silo FL for semantic parsing, there is no benchmark for this task. Thus we establish an evaluation setup by re-purposing eight single-domain text-to-SQL datasets \cite{finegan-dollak-etal-2018-improving} as eight ``clients'', which demonstrate high heterogeneity in terms of dataset sizes, domains, language usage, database structures and SQL complexity. Table \ref{data statistics} shows their statistics. 

\nop{
\begin{figure}[h]
\centering
\includegraphics[width=0.5\textwidth]{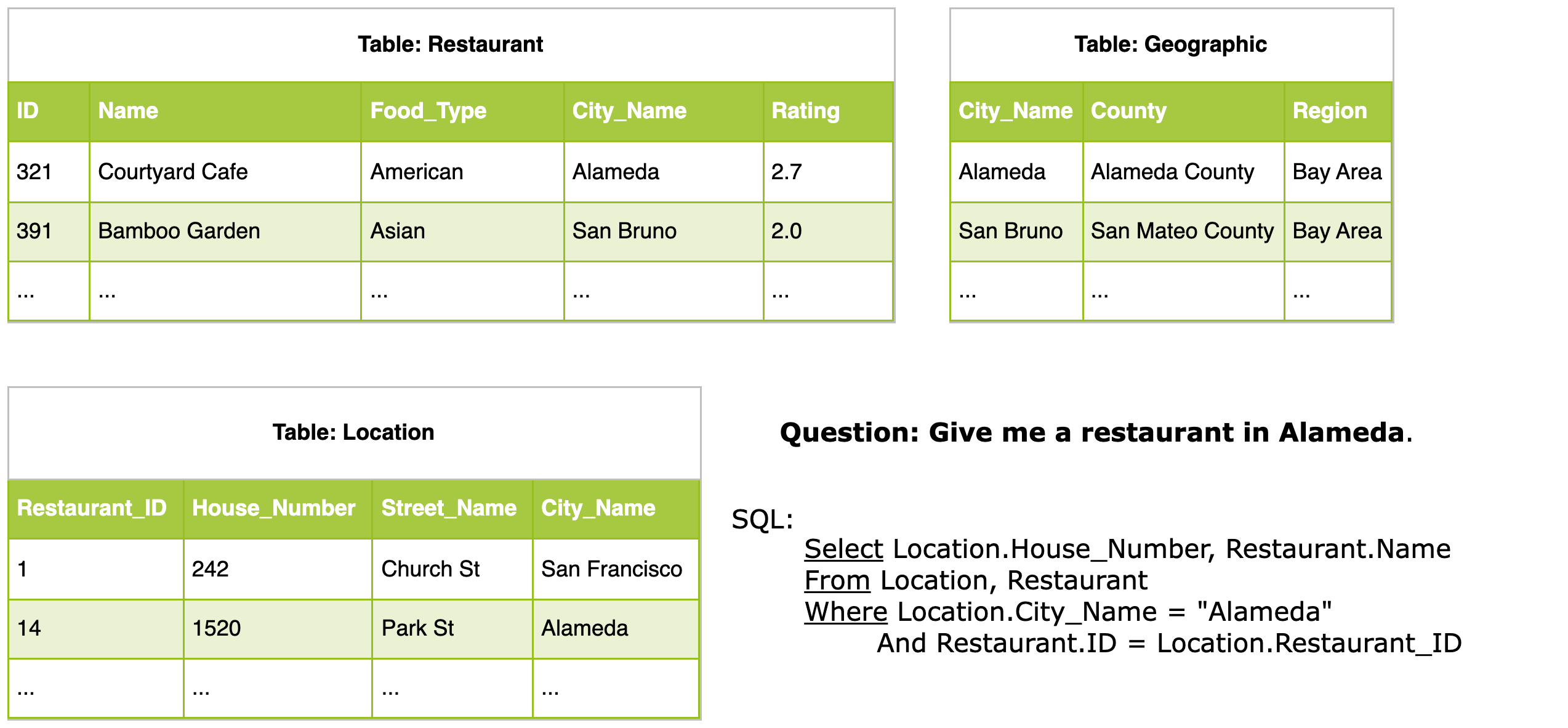}
\caption{An overview of the text-to-SQL task.}
\label{text2sql}
\end{figure}
}

Given a natural language question and the database schema, text-to-SQL parsing aims to generate a SQL query. Here the question is a sequence of tokens and the database schema consists of multiple tables with each table containing multiple columns. Figure \ref{text2sql} in Appendix shows an example of this task. We adopt T5-base as our backbone model, which has been shown as an effective unified model for various semantic parsing tasks \cite{xie-etal-2022-unifiedskg}. Similarly as in previous work \cite{xie-etal-2022-unifiedskg}, we concatenate the question tokens with the serialized relational table schemas (table names and column names) as the model input and output a sequence of SQL tokens.
\nop{Text-to-SQL is a task in which the model generates the logical form (SQL) $Y$ given natural language question $Q$ and database schema $\mathcal{S=<C,T>}$ {\wl{what is C and T?}}. It aims to automatically translate the question in human language to SQL so as to query the database to get the answer to the question. Here the question $Q = q_1,q_2,...,q_{|Q|}$ is a sequence of words. Database schema $\mathcal{S}$ contains multiple tables $t_i$ with each table containing multiple columns $c_{i,j}$. We concatenate the question token $q_k$ with the database schema $\mathcal{S}$ as the model input \zts{not very clear now}, to generate the output SQL token $y_1, y_2,...,y_{|Y|}$. Figure \ref{text2sql} shows an overview of the text-to-SQL task.\nop{\hs{Suggest to move your text-to-SQL definition from Section 2 to here. This is a better place to introduce the specifics of text-to-SQL. Section 2 is more like for semantic parsing and why FL for SP.}In Table \ref{data statistics}, Advising, ATIS, GeoQuery, Restaurants and Scholar are human-generated from NLP community, while Academic, IMDB and Yelp are human-generated from DB community} \nop{\hs{does generation by which community matter? you can still use this sentence but can you also refer to the language in intro?}. \st{Also, datasets for text-to-SQL tasks could be very complex or very simple.}}
}

The heterogeneity of the eight clients is described in detail from the following perspectives.

\textbf{Domain:} The clients are from diverse domains. Some clients such as Scholar and Academic are from closer domains than others.

\textbf{Dataset Size:} The clients differ significantly in terms of dataset sizes. Here, we consider datasets with more than 1000 train examples as \emph{large-sized} datasets, with 200$\sim$1000 as \emph{medium-sized} datasets, and with less than 200 as \emph{small-sized} datasets. In our setup, we have 2 large-sized clients (Advising and ATIS), 3 medium-sized clients (Geoquery, Restaurants and Scholar), and 3 small-sized clients (Academic, IMDB and Yelp). \nop{If we treat clients  that have more than 1000 train data as large-size clients, from 200 to 1000 as medium-size clients, and less than 200 as small-size clients, Advising and ATIS are two large-size datasets; Geoquery, Restaurants and Scholar are three medium-size datasets; and Academic, IMDB and Yelp are two small-size datasets.}

\textbf{Diversity:}
``SQL pattern count'' shows the number of SQL patterns in the full dataset. The patterns are abstracted from the SQL queries with specific table names, column names and variables anonymized. \nop{\cl{(I am not sure why we emphasize 'anonymized' here? Is it particularly related to Diversity or just general pre-processing step? Considering that anonymized has appeared in Redundancy and Complexity as well, we should carefully think about this question. If this a general pre-processing step, we only need to mention it for once at the beginning of this Section.)}\zts{It's particularly related with Diversity. The patterns also anonymize table and columns in the query, while the other two only anonymize the values}.} A larger value under this measure indicates greater diversity. In our benchmark, Advising, ATIS and Scholar have larger diversity than the other datasets.

\textbf{Redundancy:} ``Questions per unique SQL query'' counts how many natural language questions can be translated into the same SQL query (where variables are anonymized). A larger value indicates higher redundancy in the dataset. Intuitively, the higher the redundancy, the more easily a model can make correct predictions. In our benchmark, the redundancy for Advising and Restaurants is higher than the other datasets.

\textbf{Complexity:}
``Unique tables per SQL query'' (where variables in the SQL query are anonymized) represents how many unique tables are mentioned in one query. ``\texttt{SELECT}s per query'' counts how many \texttt{SELECT} clauses are included in one query. The larger these two measures, the more complex the dataset is and the more difficult for a model to make predictions. In our benchmark, Advising and ATIS are more complex. 

%% file: background-v2.tex
\section{FL for Semantic Parsing}
In this section, we first introduce the background of FL, more specifically, its training objective, training procedure and three widely adopted FL algorithms. Then we describe the motivating insights and details of our proposed mechanism.

\nop{Federated Learning is a training paradigm, which multiple clients collaboratively train a model without sharing their data. There are two realistic types of federated learning: cross-silo setting, and cross-device setting. For the cross-silo setting, clients are large institutions, such as hospitals and companies. In general, they have large computational resources and storage to train and store a large model. And it can also tolerate large communication costs between the server and clients. The number of clients is limited in this setting. For the cross-device setting, clients are small devices like mobile phones and Raspberry Pis. They have limited computational resources and storage and only small communication costs between the server and clients are affordable. It can be a huge number of clients. In our work, we study the cross-silo setting, where each client is a silo that owns its database in its special domain.
}

\subsection{Background}

\noindent \textbf{Training Objective.} Federated learning aims to optimize the following objective function:
\nop{{\cl{change the first '=' to ':=' or '= min'. Also, should we directly use Eq. (5) instead?}}}
\begin{equation}
\begin{aligned}
 \nop{\mathcal{F}(w)= \min\limits_{w}\mathbbm{E}_{i\sim \mathcal{P}}[\mathcal{L}_i(w)],\\}
 \min\limits_{w}\mathcal{F}(w) :=  \sum\nolimits_{i=1}^N p_i \mathcal{L}_i (w)\\
where \quad \mathcal{L}_i(w)= \mathbbm{E}_{b\sim \mathcal{D}_i}[f_i(w,b)].
\label{eq1}
\end{aligned}
\end{equation}

\nop{{\cl{We should have $\min\limits_{w}$ before $\mathcal{F}(w)$ in the above equation, for the second $\min\limits_{w}$ we can keep it or change $=$ to $:=$}}}

In Eqn.~\eqref{eq1}, $\mathcal{L}_i (w)$ denotes the local training objective function of the client $i$ and $N$ denotes the number of clients. $w \in \mathbbm{R}^d$ represents the parameters of the global model\nop{, and $\mathcal{P}$ denotes the distribution on the collection of clients $\mathcal{C}$}. \nop{\hs{you already used this symbol earlier? do you need it here? `distribution of clients' or `distribution of data on the clients?'}\zts{I don't think I used this symbol earlier. And the expression I use here is fine, I think. They are the distribution of clients}}$b$ denotes each batch of data. The local training loss function $f_i(w,b)$ is often the same across all the clients, while $\mathcal{D}_i$ denotes the distribution of the local client data, which is often different across the clients, capturing the heterogeneity. $p_i$ is defined as the training size proportion in Eqn. \eqref{size proportion}, where $\left|\mathcal{D}_i\right|$ is the training size of client $i$.
\begin{align}
p_i = \left|\mathcal{D}_i\right|/\sum\nolimits_{i=1}^N \left|\mathcal{D}_i\right|
\label {size proportion}   
\end{align}

\begin{figure}[t]
\centering
\includegraphics[width=0.48\textwidth]{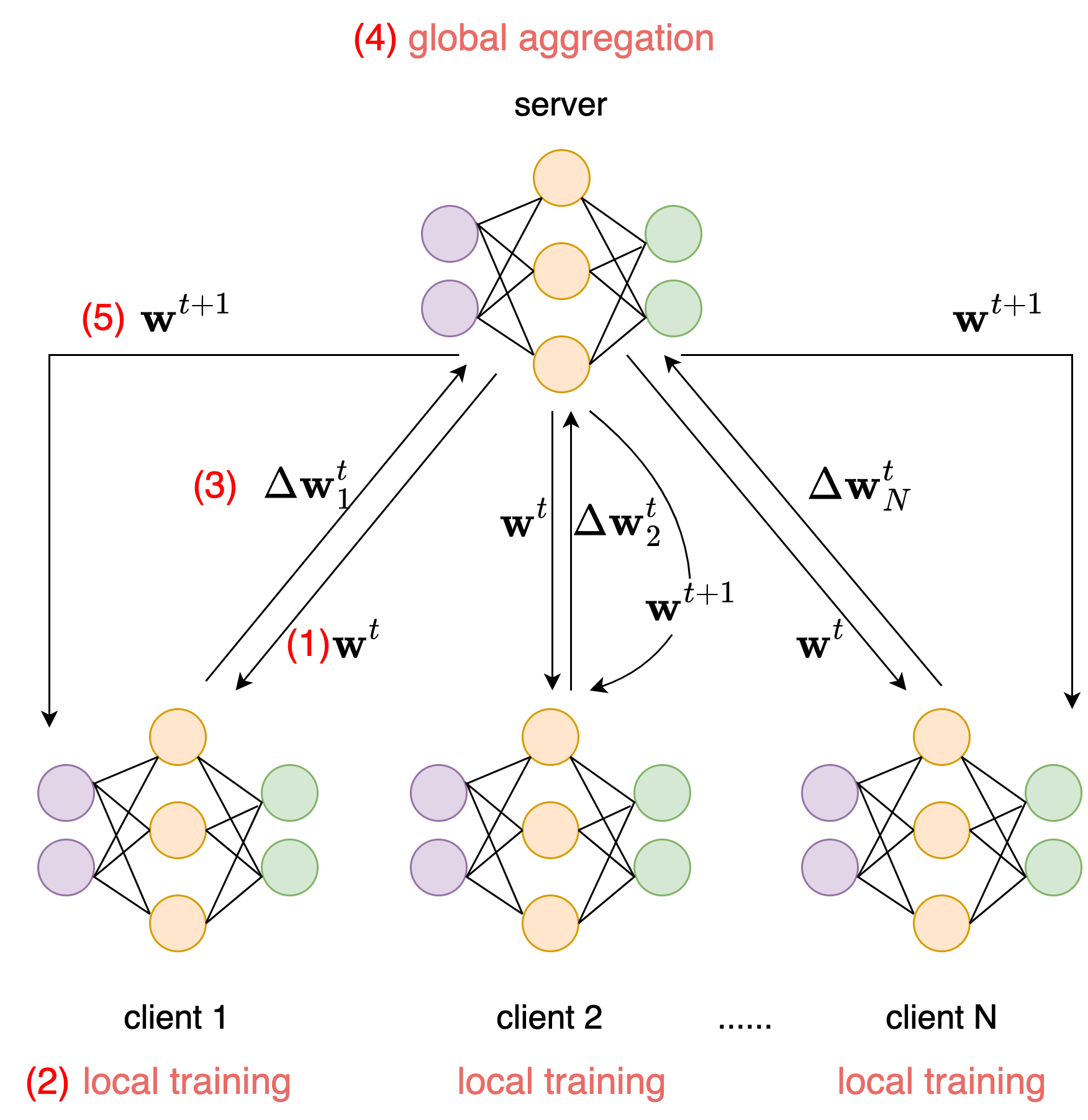}
\caption{An overview of the FL procedure. \nop{\hs{can you add `...' between the last two clients? I think you should still have `client 1' ... `client N' in the figure, maybe above `local training'? }For (4), can you say `aggregation of $\Delta w_i^t$ and global model update?'\zts{it's correct, (4) is for local training stage}}}
\label{illustration of FL}
\end{figure}

\noindent \textbf{Training Procedure.} Federated learning is an iterative process shown in Figure \ref{illustration of FL}. The server initializes the global model, followed by multiple communication rounds between the server and clients. In each \textit{communication round}, there are four steps between the server and clients. 1) In round $t$, the server sends the global model $w^t$ to all the clients. 2) After clients receive the global model $w^t$ as \nop{the locally initialized model} the initialization of the local model, they start to train it using their own data for multiple epochs and obtain the local model changes\nop{\hs{what does `accumulated' mean? accumulated over what?}} $\Delta w_i^t$ during the local training stage. 3) The clients send their local model changes to the server. 4) The server aggregates the local model changes $\Delta w_i^t$ collected from different clients as \nop{\st{the formula }}Eqn.~\eqref{aggregate gradients} shows, and then uses the $t\text{-}th$ round's global model $w^t$ and the aggregated local model changes $\Delta w^t$ to update the global model. As Eqn. \eqref{global model update} shows, $w^{t+1}$ is the global model after the update. Here, $\eta$ denotes the server learning rate. The server will send the updated model $w^{t+1}$ to the clients, then the $\text{(}t\text{+}1\text{)}\text{-}th$ round starts. 

The above procedure will repeat until the algorithm converges.
\begin{align}
\nop{\Delta w^t &= p_1 \Delta w^t_1 + p_2 \Delta w^t_2 + ... + p_N \Delta w^t_N }
\Delta w^t &= \sum\nolimits_{i=1}^N p_i \Delta w^t_i
\label{aggregate gradients} \\
w^{t+1} &= w^t - \eta\Delta w^t 
\label{global model update}
\end{align}

\noindent \textbf{FL Algorithms.} We explore three popular FL algorithms for our task:

\textit{Federated Averaging (FedAvg)} \cite{pmlr-v54-mcmahan17a} uses stochastic gradient descent (SGD) as the local training optimizer to optimize the training procedure and uses the same learning rate and the same number of local training epochs for all the clients.

\textit{FedOPT} \cite{reddi2020adaptive} is a generalized version of FedAvg. The algorithm is parameterized by two gradient-based optimizers: CLIENTOPT and SERVEROPT. CLIENTOPT is used to update the local models on the client side, while SERVEROPT treats the negative of aggregated local changes "$-\Delta w^t$" as a pseudo-gradient and applies it to the global model on the server side. FedOPT allows powerful adaptive optimizers on both server side and client side.

\nop{It has a special design to treat the negative aggregated gradients $-\Delta w^t$ \hs{what does this mean?}\hs{I still don't get what these two sentences mean.} on the server as a pseudo-gradient. Thus different adaptive optimizers can be used to update the server due to the benefit of such a pseudo-gradient. For the client side, FedOPT also allows for more powerful adaptive optimizers to optimize the local training stage.}

\textit{FedProx} \cite{li2020federated} tries to tackle the statistical heterogeneity issue by adding an L2 regularization term, which constrains the local model to be closer to the \nop{\st{initialized}}local model initialization (i.e., the global model) during each round for stable training. 

To summarize, for the local training stage, both FedAvg and FedOPT optimize the local training objective $f_i(w,b)$;
for FedProx,\nop{instead of only optimizing $f_i(w,b)$,} it optimizes Eqn.~\eqref{fedprox}, where $\mu$ is a hyperparameter and $w^t$ is the local model initialization (i.e., the global model) during the $t \text{-}th$ round. \nop{Note $w^t$ varies during each round.\hs{is this sentence necessary? Isn't it obvious? $w^t$ has a $t$ there.}}
\begin{align}
 \min\limits_{w}h_i(w,b,w^t) := f_i(w,b)+\frac{\mu}{2}\Vert w-w^t\Vert^2 
\label{fedprox}
\end{align}

For the cross-silo setting where all clients participate in training for each round, these three algorithms optimize Eqn. \eqref{eq1} during the FL process.\nop{which is equivalent to Eqn. \eqref{equivalent} \nop{\hs{how about `Eqn. \eqref{eq1} will be equivalent to Eqn. \eqref{equivalent}'}}:

\begin{align}
 \mathcal{F}(w) = \min\limits_{w} \sum\nolimits_{i=1}^N p_i \mathcal{L}_i (w)
\label {equivalent}   
\end{align}
}

%% file: proposed_method-v1.tex
\subsection{Our Proposed Re-weighting Mechanism}
\begin{figure}[t]
\begin{minipage}[t]{0.45\linewidth}
\centering
\includegraphics[width=1.45in]{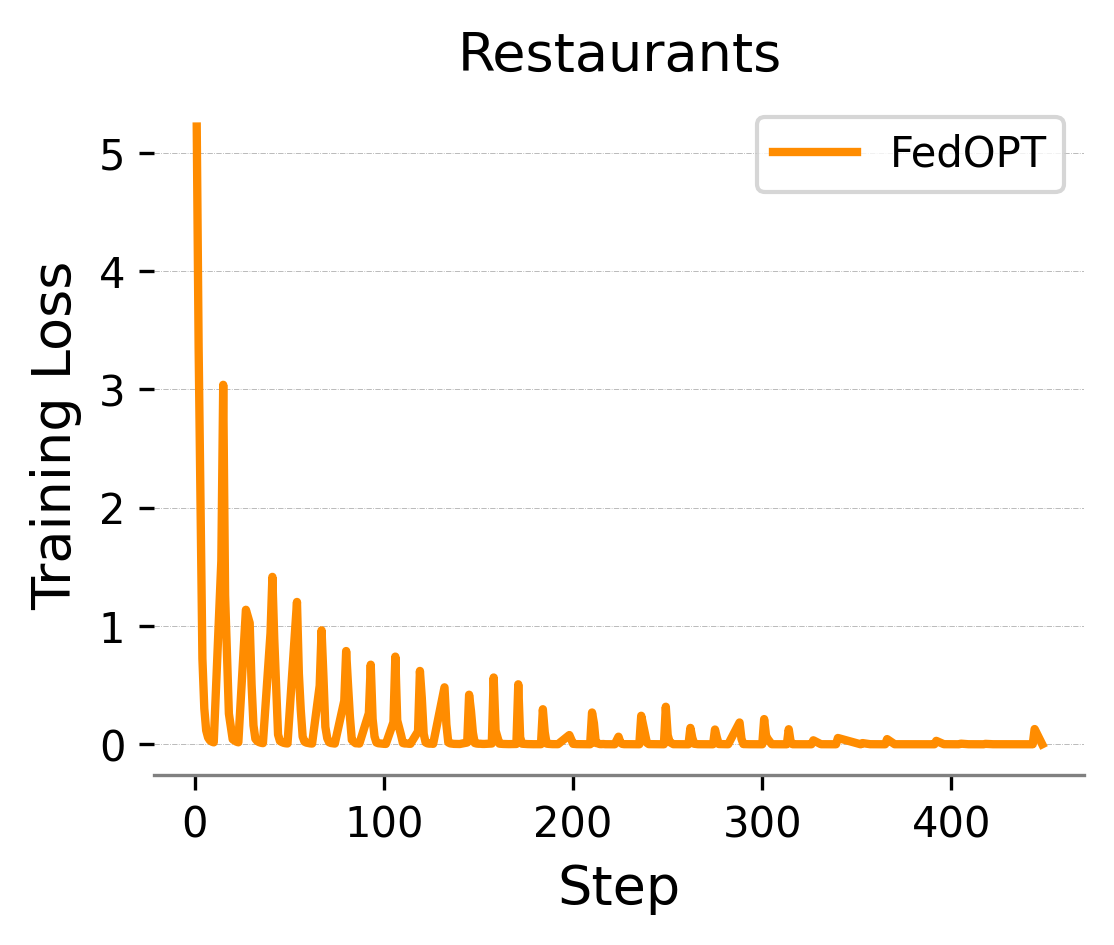}
\end{minipage}
\begin{minipage}[t]{0.45\linewidth}
\centering
\includegraphics[width=1.45in]{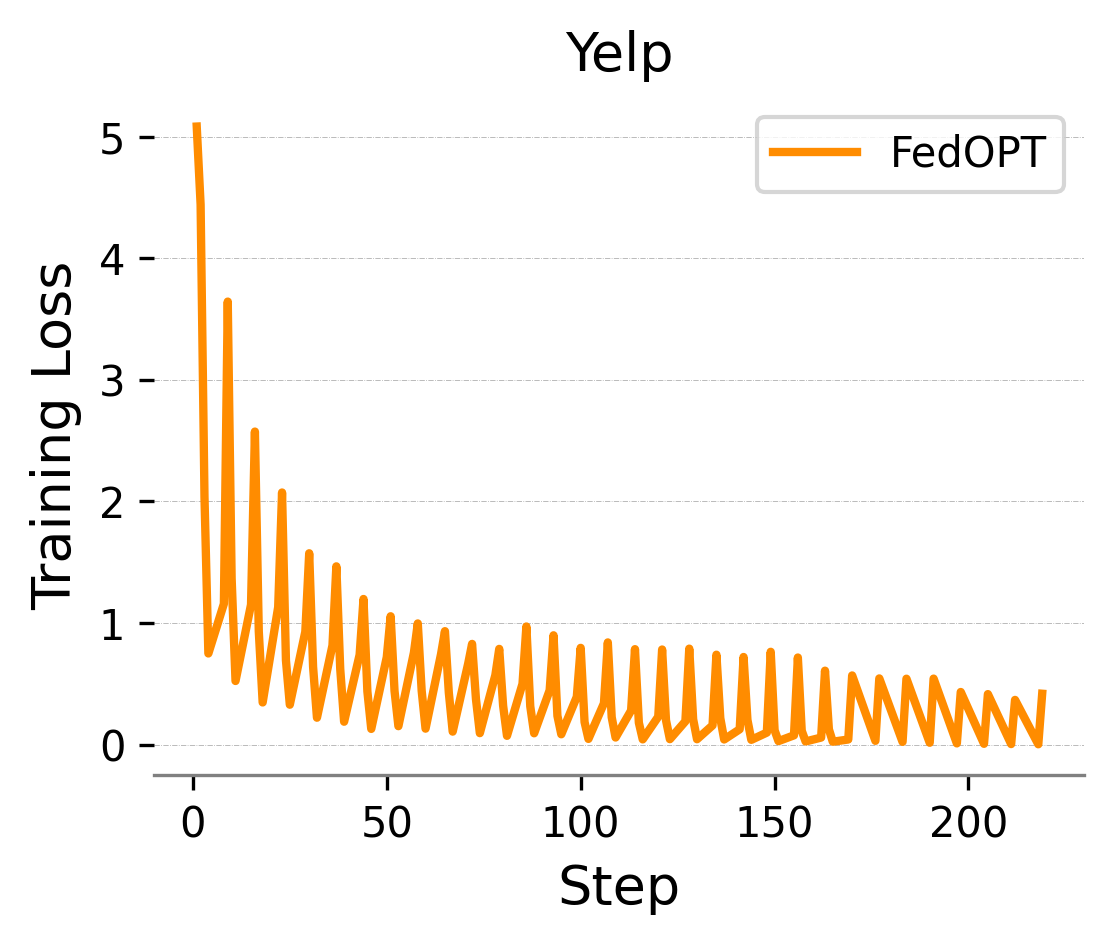}
\end{minipage}
\caption{Training loss variation of two clients: Restaurants and Yelp. The reduction of training loss during each round can signalize how far the global model is away from the client's local optimum. \nop{\hs{this is not precise from ML perspective. `fluctuation' means stable or not.}}}
\label{insight2}
\end{figure}

\noindent \textbf{Motivating Insights.} Heterogeneity, where the data distributions and dataset sizes on different clients are different, is recognized as one of the biggest challenges in FL, which usually leads to performance degradation for clients.\nop{ In our work, we can also observe the client heterogeneity from their training loss perspective} Here, we uniquely observe the clients' heterogeneity from the perspective of their training loss reduction. Take Restaurants and Yelp as two example clients. Figure \ref{insight2}\nop{, their training loss demonstrates high heterogeneity of these two clients. The figure} shows their training loss variation w.r.t. "Step". Here the "Step" is the number of iteration steps for each client during training.\nop{\st{Each down and up}} Adjacent high and low points in the figure correspond to one communication round. When the curve goes down, it means the client is in the local training stage. When the curve goes up, it means the server has updated the global model based on the aggregated local model changes from all clients and each client starts a new round of local training with the updated global model as the local model initialization. Since for different clients, the dataset sizes and the local training epochs are different, for the same communication round, the "Step" for different clients is different.\nop{\hs{I think the last sentence can be removed.}\hs{what do you mean by different? it is just an iteration step, right? you mean `the number of Steps in each round is different?'}}

As we can see, after each round, the global model deviates from the optimization trajectory \nop{\zts{obtained within the local training epochs we set).}}\nop{\hs{how about `optimization trajectory?'}\zts{It's hard to define optimization trajectory}} of each client. Thus the reduction of the training loss can signalize how far the global model is away from the client's local optimum. As $p_i$ decides how much each client contributes to the global model update, we give larger weights to those clients who have larger training loss reduction to make the global model update accommodate them better, thus mitigating potential performance degradation caused by high heterogeneity.

\nop{\noindent \textbf{Motivating Insights.} Our proposed algorithm is motivated by two key insights. The first is, we note that $p_i$ decides how much each client contributes to the global model update. The second is, the training loss of each client during the entire training process  can reflect how good the global model is for each client. \hs{then do we call these `insights?' I feel they are obvious, if we spell them out in this way.} \hs{you have to describe the setup of this figure first.} {\cl{We should blame more on the Heterogeneity in FL instead of only claiming the above insights. Specifically, we can first re-emphasize challenges caused by Heterogeneity in FL, then we take a closer look at Figure \ref{insight2} to observe heterogeneity from the local training loss perspective, then motivate our algorithm}} As Figure \ref{insight2} shows, the training loss for different clients is different during the entire training procedure. In the figure, the loss goes down means the client is in the local training stage. When the loss goes up, it means the server has updated the global model using collected gradients and distributed the new updated model to the client. So the training loss rises up at this point, which also means the global model deviates from the local optimal (suboptimal) model of the client. Thus each down and up of the training loss represents one communication round for this client. \hs{what is the relationship between `communication round' and `step' in the figure. you only mentioned the former so far.}
}

\noindent \textbf{Proposed Mechanism.} Based on the above insights, we use the training loss reduction to adjust the weight of each client, so as to reschedule its contribution to the global model update. The final weight is formulated as Eqn. (\ref{simplified_weight}), where $\Delta \mathcal{L}_i^t$ is the training loss reduction during the $t\text{-}th$ round.\nop{We introduce the loss reduction proportion in Eqn.~\eqref{weight_loss} to help adjust the weight during each round $t$.
\nop{\hs{can we directly say `loss reduction $l_i^t$'}  Note $m_i$ varies for each round $t$ {\wl{let's revise the following equations based on our previous discussion}}\zts{I'm thinking to put the simplified formula to the final discussion section to mention that we don't need to transmit the loss to the server. I feel since the formula is simple, we should put the formulation process here rather than directly put a simplified version here} \hs{I think you should just put the simplified version here.}.}
\begin{align}
s_i = \left|\mathcal{D}_i\right|/\sum\nolimits_{i=1}^N \left|\mathcal{D}_i\right|
\label{weight_size}\\
m_i^t = \Delta \mathcal{L}_i^t / \sum\nolimits_{i \in C_t} \Delta l_i^t 
\label{weight_loss}\\
p_i^t = s_i m_i^t / \sum\nolimits_{i \in C_t} s_i m_i^t 
\label{weight}
\end{align}
The above process can be simplified as Eqn. (\ref{simplified_weight}). Compared with FedAvg, FedOPT and FedProx, we use the training loss reduction to adjust the weight of each client's contribution to the global model update.}
\begin{align}
\nop{
s_i = 
\left|\mathcal{D}_i\right|/\sum\nolimits_{i \in C_t} \left|\mathcal{D}_i\right|
\label{weight_size}\\}
p_i^t = \left|\mathcal{D}_i\right| \Delta \mathcal{L}_i^t / \sum\nolimits_{i=1}^N
\left|\mathcal{D}_i\right| \Delta \mathcal{L}_i^t 
\label{simplified_weight}
\end{align}
\nop{\cl{Just realized that we use small $\Delta l_i^t$ instead of Capital $L$ here. Combining with Line 21, should we change to captal $\mathcal{L}$? Otherwise we may want to define what is $\Delta l_i^t$ }}
\nop{\hs{also Figure 1 (where we used $p$ and $p'$) should be consistent: maybe change $p'$ and $r_i$ there to be associated with $t$.} }

\nop{
FedAvg, FedOPT and FedProx all use the training size proportion shown in \eqref{weight_size} as the weight. Though they try to improve federated learning performance from different perspectives, they ignore the importance of weight. \hs{Some parts of this paragraph are already said earlier. I think here you could directly say compared with them, we use the loss reduction to adjust their weight, and we don't need to transmit additional parameters.} Instead, we add another factor \eqref{weight_loss}: the loss reduction proportion term to help automatically control the contribution of each client during the aggregation. We then multiply both the training size proportion $s_i$ and the loss reduction proportion $m_i$ and normalize it to get the final weight $p_i$ (as shown in \eqref{weight}).
}
FedAvg, FedOPT, FedreProx with our proposed mechanism\nop{and our proposed algorithm} are summarized in Algorithm \ref{algorithm} in Appendix.

%% file: experiments-v1.tex
\section{Experiments}
\nop{{\cl{Let's plan to write a summary of experimental results so that reviewer can quickly get our key observations. It could be something like: we aim to use XXX datasets to verify effectiveness of our method and we observe that our method achieves better performance....}} \zts{do you mean we can put the same takeaways as abstract here?}\hs{could be a bit more detailed than abstract, but some take-aways from this section.} {\cl{I see we now seperate experimental results to a separate Section 6, then we may need to summarize the key takaways at the beginning of Section 6}}}

\noindent \textbf{Datasets.} We re-purpose eight datasets: ATIS \citep{data-atis-geography-scholar, data-atis-original}, GeoQuery \citep{data-atis-geography-scholar, zelle1996learning}, Restaurants \citep{data-restaurants-logic, data-restaurants-original, data-restaurants}, Scholar \citep{data-atis-geography-scholar}, Academic \citep{data-academic}, Advising \citep{finegan-dollak-etal-2018-improving}, Yelp and IMDB \citep{data-sql-imdb-yelp} as eight clients. These datasets have been standardized to the same SQL style by \citet{finegan-dollak-etal-2018-improving}. Their characteristics have been described in Section \ref{benchmark}. We follow "question split" datasets preprocessed by \citet{finegan-dollak-etal-2018-improving} to split the train, dev and test data, which means we let the train, dev and test examples have different questions but the same SQL queries are allowed\nop{\cl{do we really need these detailed information? If we follow previous work, probably not. If such informaiton is essential, we can change to: where the train, dev and test examples have different questions and may have overlapping queries.}}. For Advising, ATIS, GeoQuery and Scholar, we directly use the original question split as our split. For Restaurants, Academic, IMDB and Yelp, since the data sizes are relatively small, the original question split uses 10 splits for cross validation without specifying train, dev and test examples.\nop{To simplify the comparison among the individual finetuning, centralized and FL experimental settings} Given FL is costly as we need multiple GPUs to finish one experiment, we fix the train, dev and test set by randomly selecting 6 splits as the train set, 2 splits as the dev set and 2 splits as the test set. 

\noindent \textbf{Evaluation Metrics.}
1) Exact Match (EM): a prediction is deemed correct only if it is exactly the same as the ground truth (i.e., exact string match), which is widely used for text-to-SQL parsing~\cite{finegan-dollak-etal-2018-improving}. All the evaluations in our experiments consider the values generated in the SQL query\nop{{\cl{this sentence is hard to understand, what is values here, what is the output here? do we really need this sentence?}}}. 2) MacroAvg: The arithmetic mean of EM across all clients, which treats each client equally. 3) MicroAvg: The total number of correct predictions on all the clients divided by the total test examples, which treats each test example equally.

\noindent \textbf{Learning Paradigm.} We compare three learning paradigms: finetuning, centralized and FL. 1) \textit{Finetuning}: we individually finetune our backbone model (T5-base) on the training data of each client. 2) \textit{Centralized}: we merge the training data of all the clients and finetune our backbone model on the merged training data to obtain one model. 3) \textit{FL}: we leverage eight clients and a server to learn a global model without sharing each client's local data. By comparing individual finetuning and FL, we can show the benefit of FL for some clients, especially for small-sized clients.\nop{The centralized paradigm is less practical compared with the other two paradigms but can serve as an upper bound for FL, which can show how large the room is for FL to fully utilize the training data from all the clients } The centralized paradigm is less practical compared with the other two paradigms due to privacy considerations. However, it can serve as a useful reference to help validate how effective an FL algorithm is in fully exploiting heterogeneous data across multiple clients. 

\noindent \textbf{Implementation Details.}
We implement the FL algorithms and T5-base model\nop{develop our experiments} based on FedNLP \cite{lin-etal-2022-fednlp}, FedML \cite{chaoyanghe2020fedml} and UnifiedSKG \cite{xie-etal-2022-unifiedskg}. We use Adafactor \cite{pmlr-v80-shazeer18a} as the optimizer for finetuning and centralized paradigms, and as the client optimizer\footnote{Note we use Adafactor as the local optimizer for FedAvg, so the FedAvg in our paper is slightly different from the original proposed FedAvg, which uses stochastic gradient descent(SGD) as the local optimizer.} for FL paradigm, since it has been shown as the best optimizer to optimize the T5 model. More details are in Appendix \ref{implementation details}.

For the computing resources, we use 1 NVIDIA A6000 48GB GPU for finetuning, with batch size 8. We use 2 NVIDIA A6000 48GB GPUs for centralized training, with batch size 8. We use 5 NVIDIA A6000 48GB GPUs for all federated learning experiments. Specifically, one GPU is used as the server and the other four GPUs are used as 8 clients, with each GPU accommodating 2 clients. The batch size for clients GeoQuery, Restaurants, Scholar, Academic, IMDB and Yelp is 4, and for clients Advising and ATIS is 8.

\begin{table*}[t]
\centering
\resizebox{\linewidth}{!}{
\begin{tabular}
{@{}lcccccccccc@{}}
\toprule
\text{} & \text{Advising\dag} & \text{ATIS\dag} & \text{GeoQuery\S} & \text{Restaurants\S} & \text{Scholar\S} & \text{Academic*} & \text{IMDB*} & \text{Yelp*} & \text{MacroAvg} & \text{MicroAvg}\\
\cmidrule(lr){1-9}\cmidrule(lr){10-11}
Finetuning & 84.47 & 53.91 & 72.76 & 98.65 & 74.31 & 57.89 & 26.92 & 33.33 & 62.78 & 71.47 \\
Centralized & 85.51 & 56.38 & 79.21 & 100 & 72.48 & 65.79 & 61.54 & 41.67 & 70.32 & 74.21 \\
\midrule
FedOPT & 79.76 & 51.23 & \textbf{77.42} & \textbf{98.65} & \textbf{66.51} & 50 & 34.62 & 8.33 & 58.32 & 68.49\\
FedOPT$_{lorar}$ & \textbf{80.98} & \textbf{52.35} & 75.99 & \textbf{98.65} & 64.68 & \textbf{68.42} & \textbf{38.46} & \textbf{20.83} & \textbf{62.55} & \textbf{69.39}\\
\midrule
FedAvg & \textbf{76.44} & \textbf{50.11} & 59.86 & 72.97 & 38.07 & 2.63 & 7.69 & 12.5 & 40.03 & 57.89\\
FedAvg$_{lorar}$ & 74.69 & 49.89 & \textbf{68.82} & \textbf{98.65} & \textbf{52.29} & \textbf{65.79} & \textbf{46.15} & \textbf{25} & \textbf{60.16} & \textbf{63.91}\\
\midrule
FedProx & \textbf{74.52} & \textbf{50.56} & 65.95 & 81.08 & 38.53 & 10.53 & 3.85 & 8.33 & 41.67 & 58.84\\
FedProx$_{lorar}$ & 73.12 & 49.66 & \textbf{67.38} & \textbf{98.65} & \textbf{48.17} & \textbf{63.16} & \textbf{46.15} & \textbf{20.83} & \textbf{58.39} &\textbf{62.42}\\
\bottomrule
\end{tabular}}
\caption{\label{major results}
Main results for different learning paradigms and FL algorithms. "\dag": large-sized clients. "\S": medium-sized clients. "*": small-sized clients.\nop{avg refers to the average performance of all eight clients. wavg means the weighted average performance of all eight clients based on their test sizes. The notation "lcar" refers to our proposed algorithm. \hs{what does the `*' mean? you need to explain it.}}
}
\end{table*}

\section{Results and Analysis}

\nop{In this section, we compare finetuning, centralized and FL paradigms for both individual client performance and average performance. In addition, we compare the training loss between FedOPT and FedOPT$_{lorar}$, and provide the performance of alternative weight mechanisms. At last, we show that \texttt{Lorar} is more robust to local training epochs in Appendix \ref{random epoch section}. \zts{I didn't put the detailed conclusions here. We exceed 8 page limit, I feel this may be not important.}{\cl{I agree. We may not need this paragraph to save space.}}
}

\subsection{Main Results}\label{main results section}
\nop{{\cl{Let’s reorganize Section 7.1 to highlight key take aways easily. For example, we can break the observations into the following points: 1) Larger and more diversified data benefit model performance (centralized setting is better than finetuning: paragraph 1); 2) Federated Learning degrades performance as compared to centralized setting but protect privacy (paragraph 2); 3) Our algorithm improves baseline FL algorithm (paragraph 3) 4) A closer look at each individual client for finer-grained analysis (paragraph 4 and after)}} }

\nop{\st{{\textit{(1) Larger and more diversified data benefit model performance as centralized setting is better than finetuning.}}We first finetune T5-base on all eight datasets. And then we merge all eight datasets and finetune T5-base on the merged dataset to get the centralized training results.}} 

\noindent \textbf{Centralized vs. Finetuning.} As Table \ref{major results} shows, compared with the individual finetuning setting, the model performance under the centralized setting has been improved on all the datasets except Scholar. \textit{This means merging all the data to train a model, which increases the size and diversity of training data, can improve the model's generalization ability and lead to improvement for most datasets.} This observation also motivates us to leverage these datasets to study FL for semantic parsing, which is a more practical paradigm than the centralized one.\nop{\hs{i feel the reason is not very convincing. maybe remove the sentence:} While for Scholar, we observe that the performance drops slightly after merging other datasets, which may be due to the noise brought by them.} 

\begin{figure*}[htbp]
\subfigure{
\includegraphics[width=0.23\textwidth]{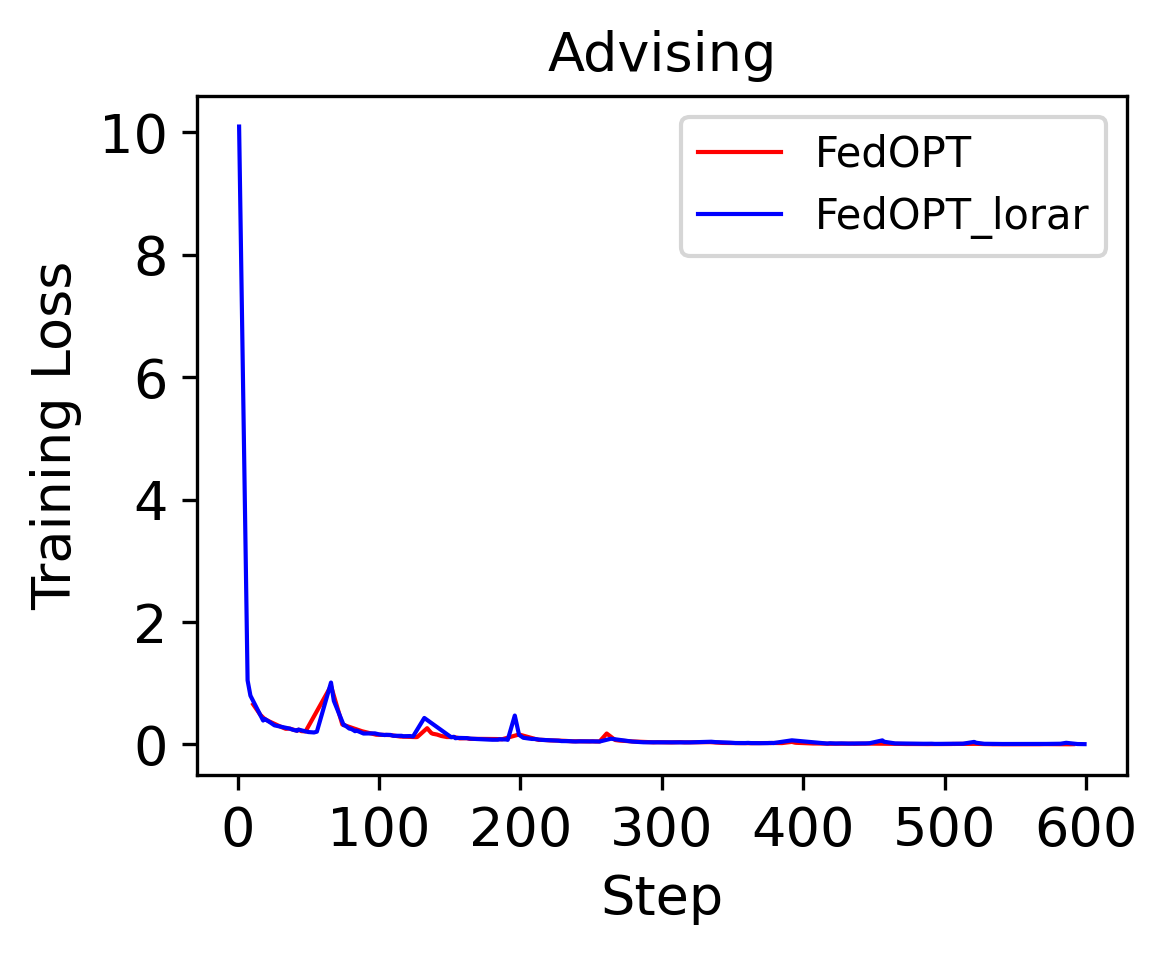}
}
\subfigure{
\includegraphics[width=0.23\textwidth]{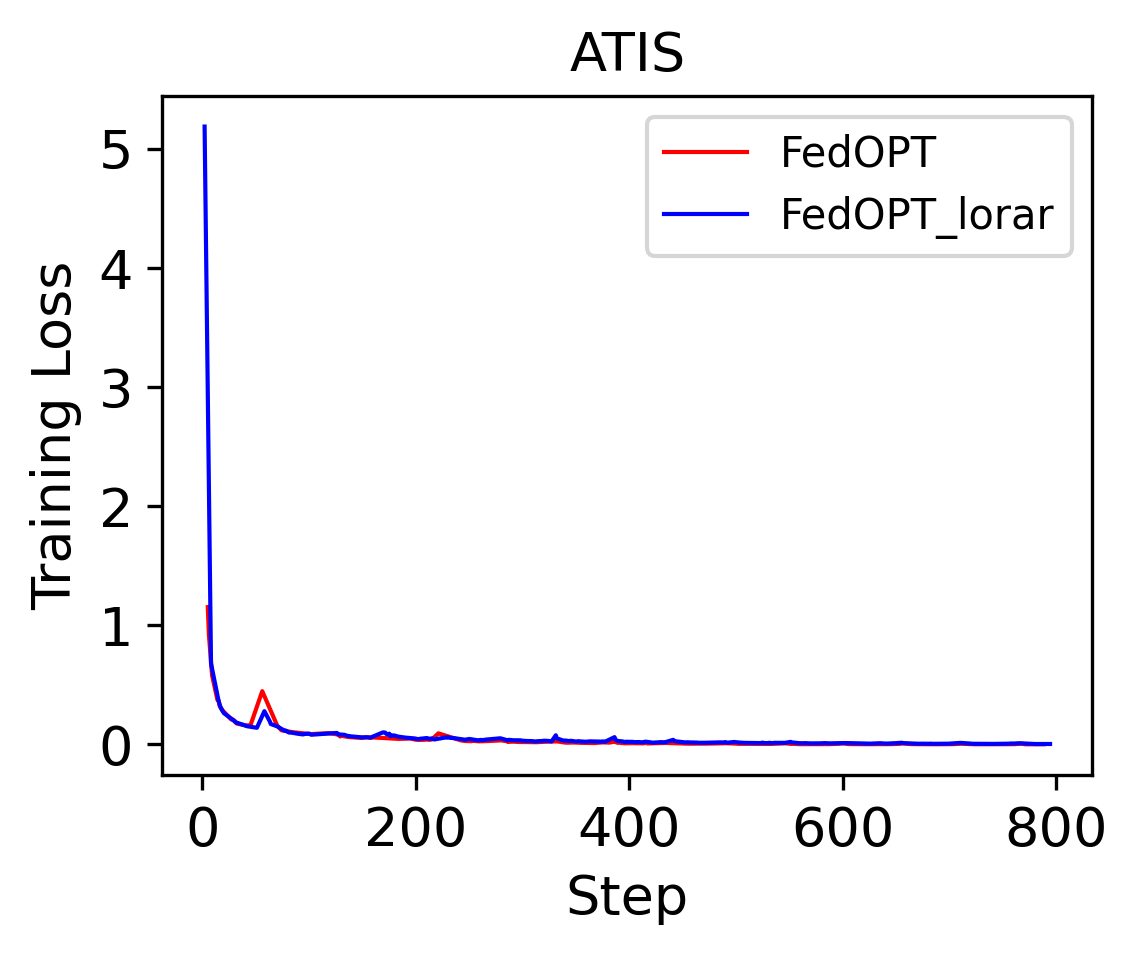}
}
\subfigure{
\includegraphics[width=0.23\textwidth]{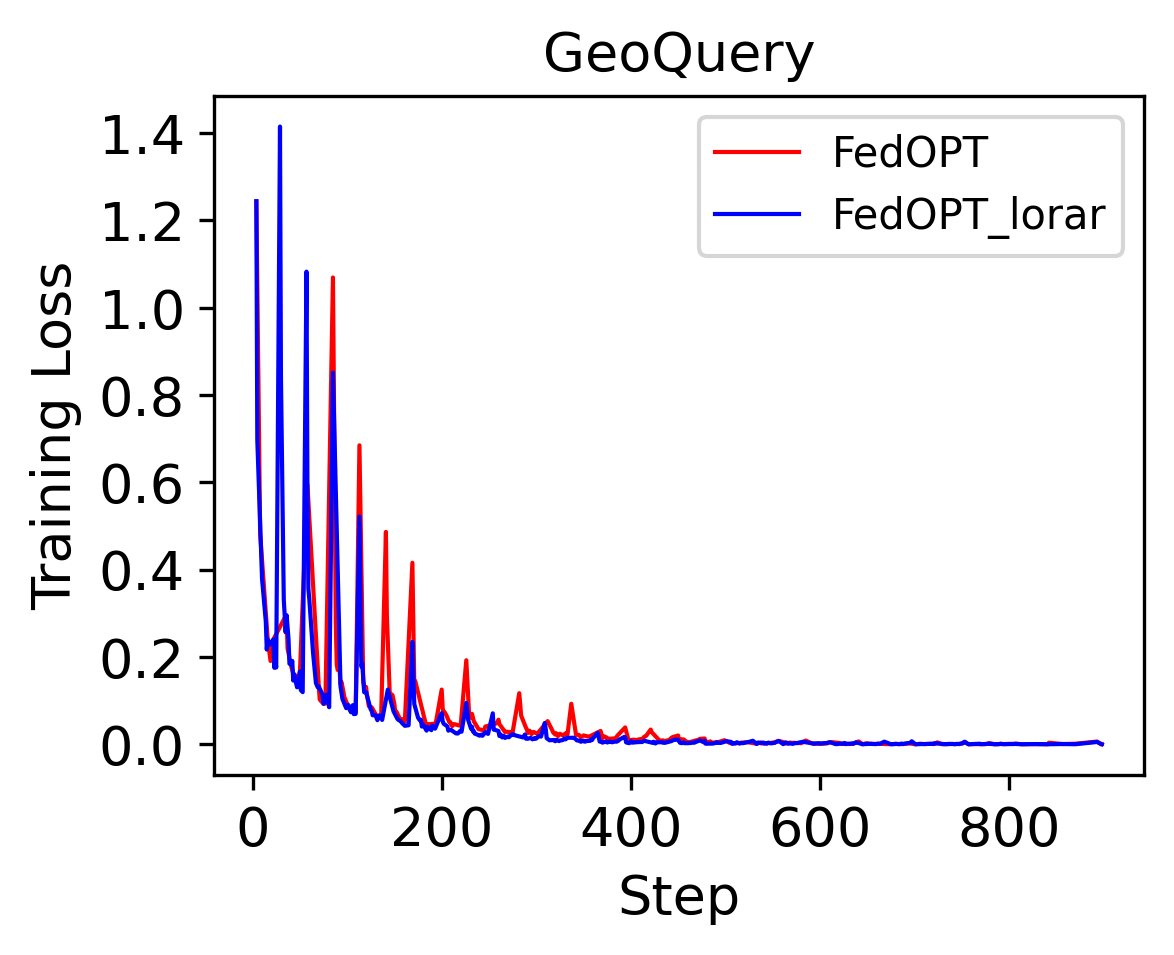}
}
\subfigure{
\includegraphics[width=0.23\textwidth]{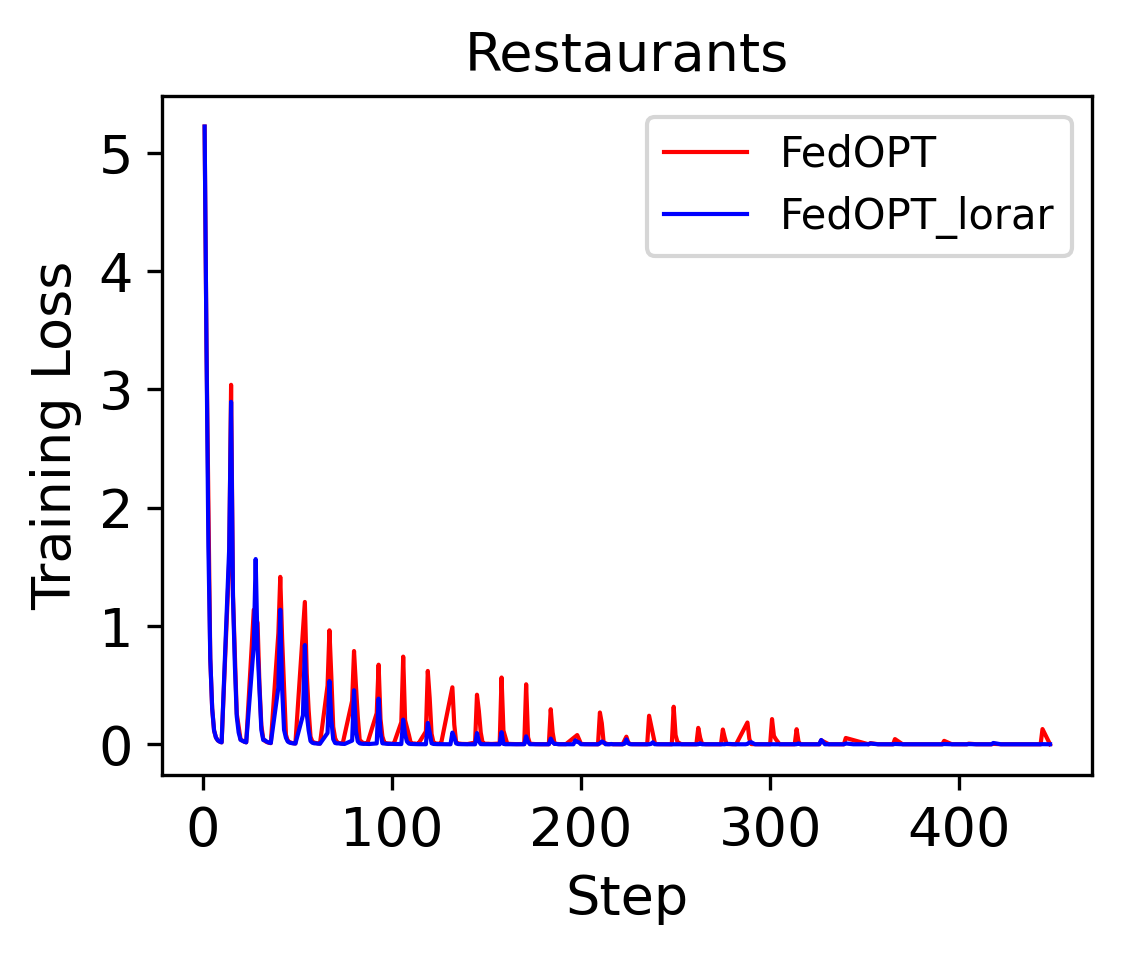}
}
\subfigure{
\includegraphics[width=0.23\textwidth]{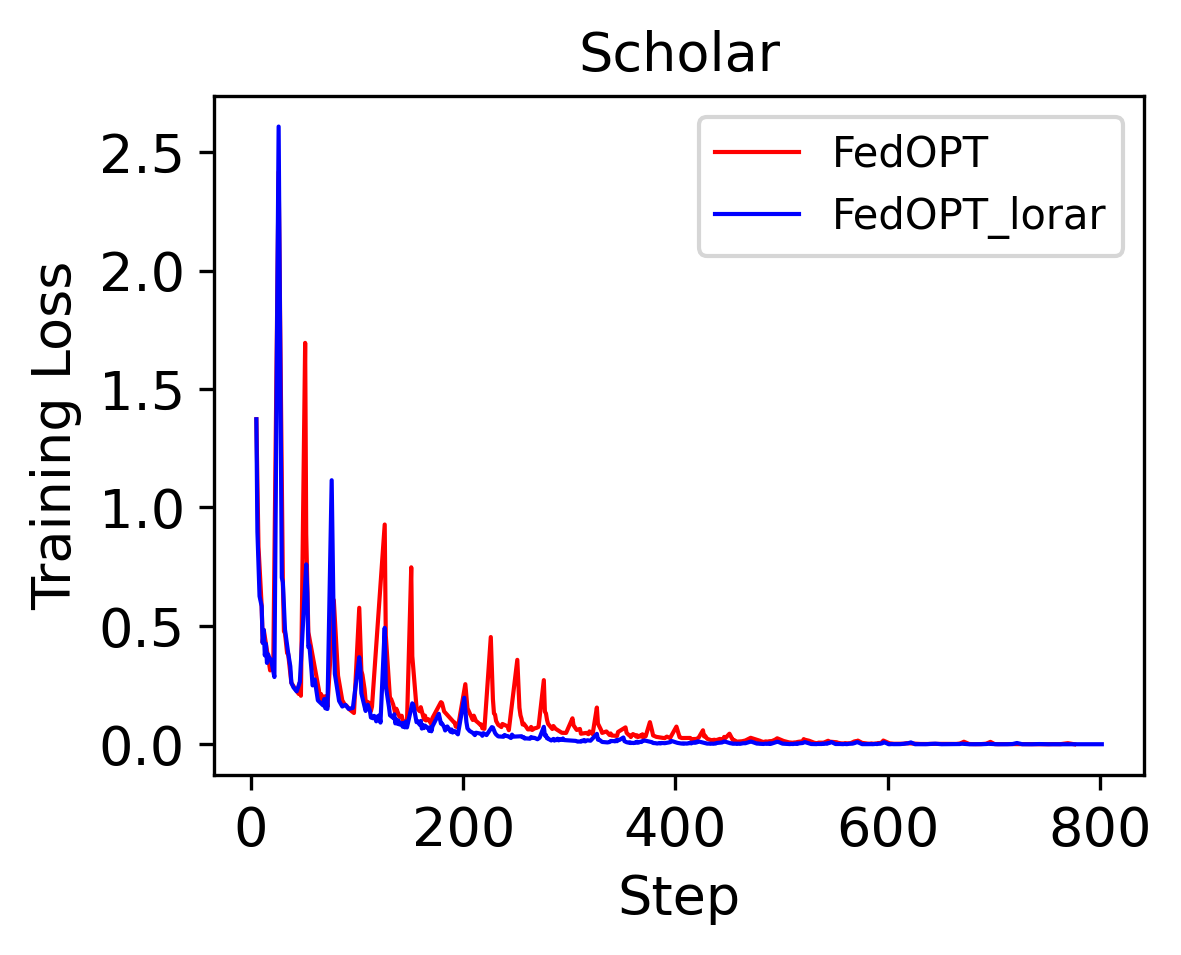}
}
\subfigure{
\includegraphics[width=0.23\textwidth]{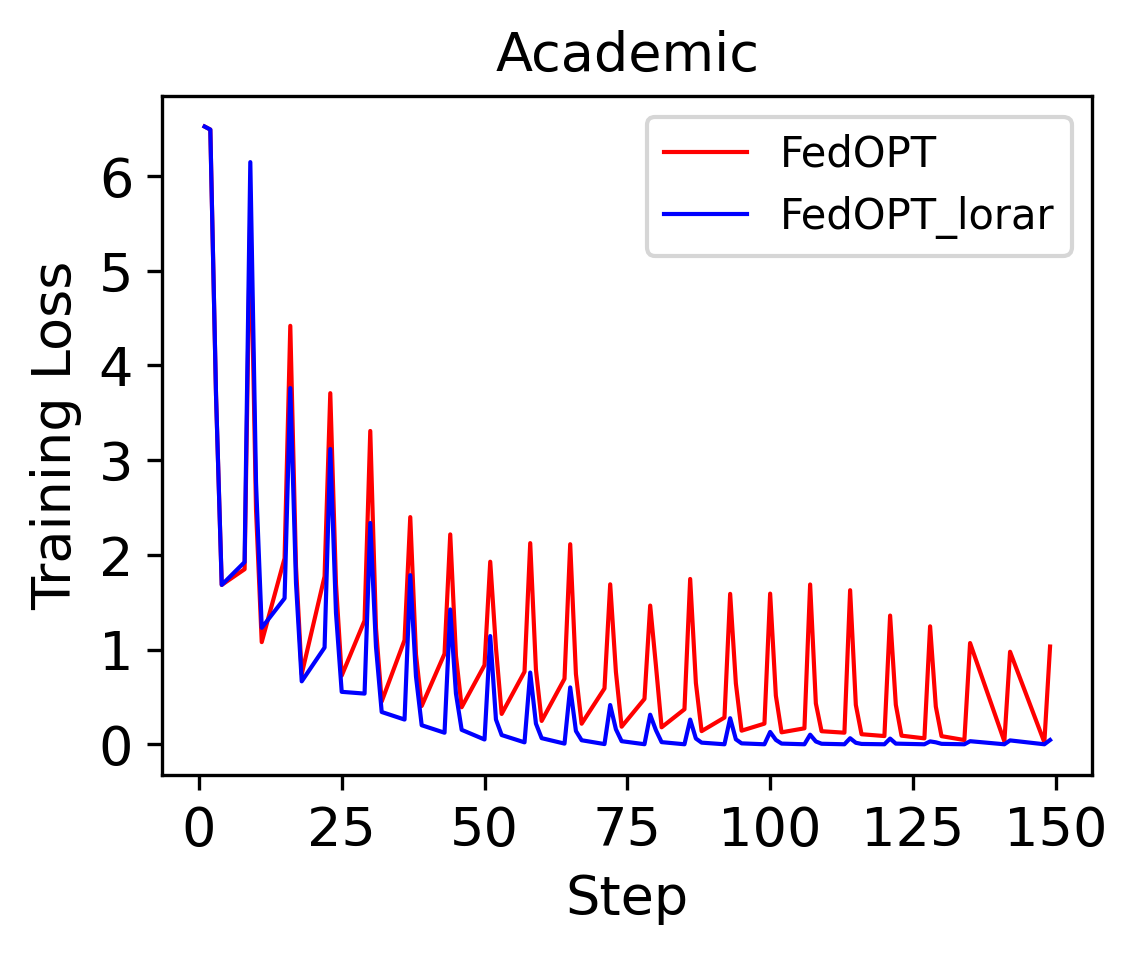}
}
\subfigure{
\includegraphics[width=0.23\textwidth]{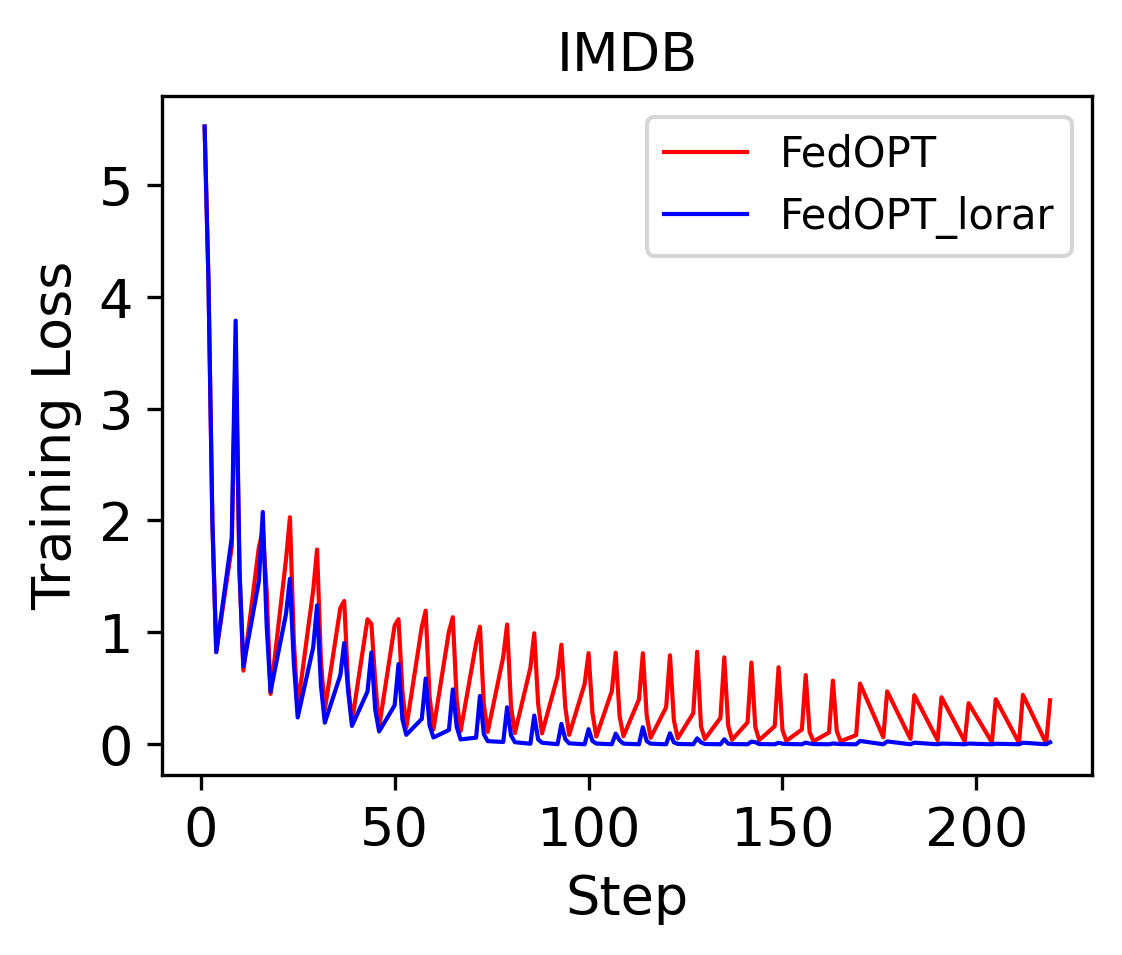}
}
\subfigure{
\includegraphics[width=0.23\textwidth]{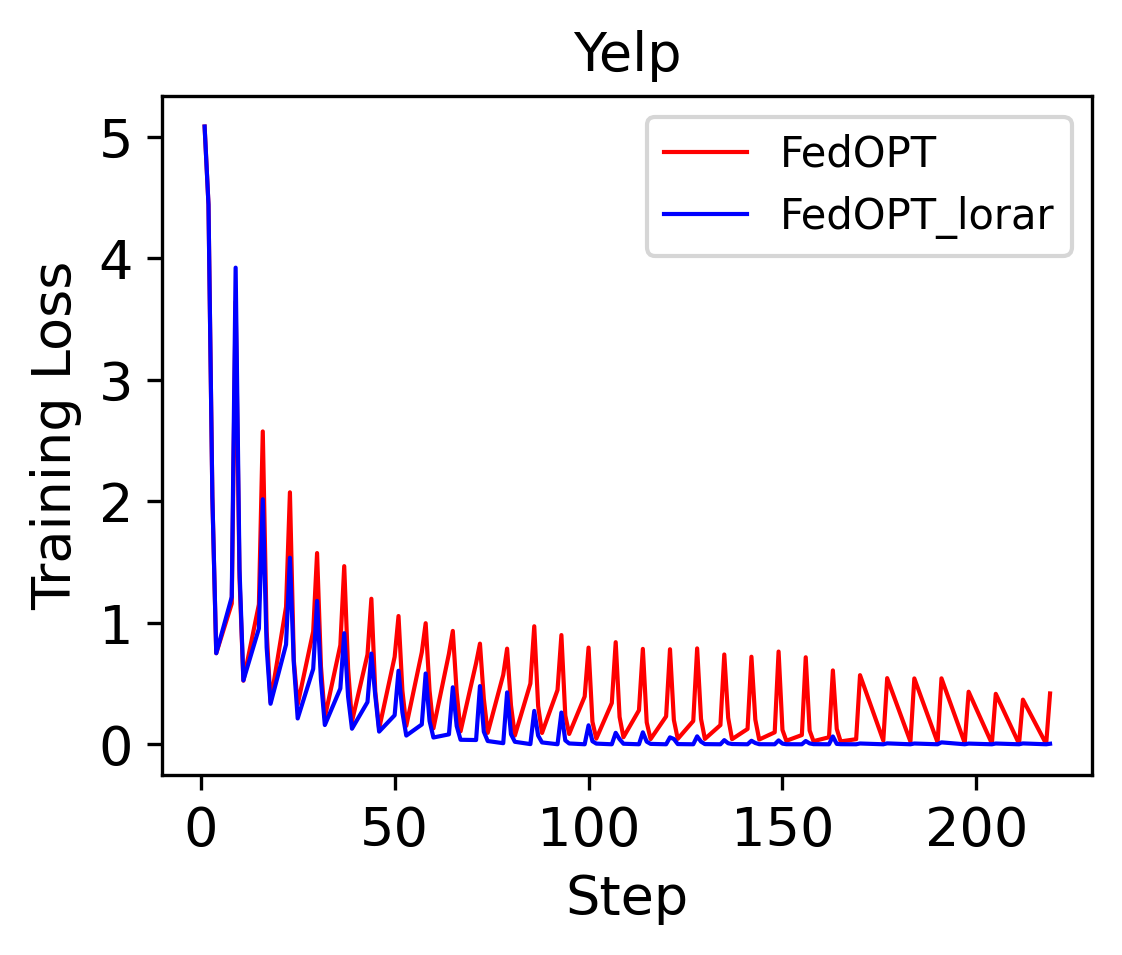}
}
\caption{Training loss variation on eight clients for FedOPT and FedOPT$_{\texttt{lorar}}$. \nop{\hs{the figures here are still blurry. note the differences between your previous ones and this one.}}}
\label{eight clients}
\end{figure*}

\noindent\textbf{Effectiveness of \texttt{Lorar} in FL.} Applying our proposed \texttt{Lorar} mechanism can substantially improve the performance of all three FL algorithms overall. 
As Table \ref{major results} shows, for FedOPT, our proposed FedOPT$_{\texttt{lorar}}$ performs substantially better or similarly on all clients, except for a slight drop on GeoQuery and Scholar. Moreover, on the three smaller datasets: Academic, IMDB and Yelp, \texttt{Lorar} brings much larger performance gains. For FedAvg and FedProx, in addition to these three datasets, \texttt{Lorar} also brings substantial improvements on two medium-sized clients: Restaurants and Scholar.\nop{This shows that our proposed mechanism is very effective overall across different FL algorithms and clients.} These observations validate the effectiveness of our proposed mechanism under different FL algorithms and across different clients.

We additionally analyze these three FL algorithms and their performance variation with and without using \texttt{Lorar} under different communication rounds. More details are included in Appendix \ref{FL baselines comparison} and \ref{performance changes}.

\noindent\textbf{FL vs. Finetuning/Centralized.}
As Table \ref{major results} shows, the original FedOPT outperforms finetuning on GeoQuery and IMDB, which shows that FL can boost the model performance for some clients. In addition, although there is still a gap between existing FL algorithms (FedOPT, FedAvg, and FedProx) and the centralized setting, by equipping them with our proposed \texttt{Lorar}, we can reduce the gap by 4-20 points (i.e., absolute difference under MacroAvg). 
It is worth noting that institutions are often reluctant or prohibited to share their data in practice, especially for SQL data that may directly reveal private database content. 
Therefore, the centralized paradigm is impractical. 
Nonetheless, it can serve as a useful reference to help validate how effective an FL algorithm is in fully exploiting heterogeneous data across multiple clients. 
The results show that our benchmark provides a challenging testbed for a realistic FL problem, and there is still a large room to further improve the FL algorithms. 

\begin{table*}[h]
\centering
\resizebox{\linewidth}{!}{
\begin{tabular}
{@{}lcccccccccc@{}}
\toprule
\text{} & \text{Advising} & \text{ATIS} & \text{GeoQuery} & \text{Restaurants} & \text{Scholar} & \text{Academic} & \text{IMDB} & \text{Yelp} & \text{MacroAvg} & \text{MicroAvg}\\
\cmidrule(lr){1-9}\cmidrule(lr){10-11} 
FedOPT & 79.76 & 51.23 & 77.42 & \textbf{98.65} & \textbf{66.51} & 50 & 34.62 & 8.33 & 58.32 & 68.49\\
FedOPT$_{lr}$ & 75.04 & \textbf{53.47} & 75.63 & \textbf{98.65} & 62.39 & 60.53 & 34.62 & \textbf{25} & 60.67 & 67.12\\
FedOPT$_{equal}$ & 76.96 & 53.02 & \textbf{77.78} & \textbf{98.65} & 63.3 & 63.16 & 34.62 & 20.83 & 61.04 & 68.13 \\
FedOPT$_{lorar}$ & \textbf{80.98} & 52.35 & 75.99 & \textbf{98.65} & 64.68 & \textbf{68.42} & \textbf{38.46} & 20.83 & \textbf{62.55} & \textbf{69.39}\\

\bottomrule
\end{tabular}}
\caption{\label{ablation study}
Alternative weighting mechanisms for FedOPT on the test set of our proposed benchmark, where FedOPT only uses a client's training set size (w/o loss reduction) as its weight, FedOPT$_{lr}$ only uses a client's loss reduction during each round (w/o train set size) as its weight, FedOPT$_{\texttt{lorar}}$ considers both factors as its weight (Eqn. (\ref{simplified_weight})) and FedOPT$_{equal}$ gives each client equal weight (w/o considering both factors).}\nop{Recall that FedOPT uses a client's training set size (w/o loss reduction) as its weight, FedOPT$_{lr}$ refers to only using a client's loss reduction during each round (w/o train set size) as its weight, while FedOPT$_{\texttt{lorar}}$ considers both factors (Eqn. (\ref{simplified_weight})). FedOPT$_{equal}$ means each client gets equal weight.}

\end{table*}

\subsection{Training Loss Analysis}
To better understand how \texttt{Lorar} affects the training process in FL, we show the training loss variation for FedOPT and FedOPT$_{\texttt{lorar}}$ in Figure~\ref{eight clients}. For FedOPT,\nop{\cl{we observe that 1) for larger datasets... 2) for smaller datasets....}} we can see for larger datasets such as Advising and ATIS, the training converges much faster and the global model is closer to the client's local optimum within very few rounds. While for smaller datasets such as Academic, IMDB and Yelp, the training loss oscillates widely, which means the global model converges slower for these clients (if at all). 
After applying \texttt{Lorar}, however, \textit{the training loss converges faster on almost all the clients}, which means the global model can get close to the client's local optimum more quickly and easily. \nop{\cl{Can be modified to sth like: In comparison, after applying \texttt{Lorar} to FedOPT, the training loss converges faster on almost all the clients meaning that the global model move closer and faster to the client's optimal local model}}

\subsection{Alternative Weighting Mechanisms}

As FedOPT performs best among all three FL baselines, we use it to compare \texttt{Lorar} with alternative weighting mechanisms. As Table \ref{ablation study} shows, \texttt{Lorar}, which considers both the training set size and the loss reduction 
in the weight, can achieve the best results. Comparing FedOPT$_{lr}$ (i.e., FedOPT with only loss reduction considered in the weight) and FedOPT$_{\texttt{lorar}}$, we can see removing the training set size from the weight will lead to a large drop under MacroAvg and MicroAvg, which indicates that training set size is an important factor during the aggregation. This is intuitive since for those clients which have more training data, their local models tend to be more reliable and more generalizable.\nop{Thus it's reasonable that using the loss reduction proportion to adjust the contribution based on the training size proportion can achieve the best results.}\nop{We also try giving each client the same weight (i.e., FedOPT$_{equal}$),\nop{which means during the global model update, we treat each client with the same importance. Though the performance is close to the FedOPT that uses the training size proportion as weight,} which underperforms FedOPT$_{\texttt{lorar}}$.} We also compare with FedOPT$_{equal}$ where all clients are given the same weight. We can see that our FedOPT$_{\texttt{lorar}}$ yields superior performance. The conclusion can also be verified in Figure \ref{fedopt_ablation_study} in Appendix, where we show their performance variation under different communication rounds.

\subsection{Impact from Dataset Heterogeneity}
(1) The impact of diversity, redundancy and complexity: In Table \ref{major results} and \ref{ablation study}, for Restaurants, the results of finetuning, centralized training, and varying weighting mechanisms of FedOPT are pretty close and all very high (close to 100\%), which shows it is a relatively easy dataset for any learning paradigm and weighting mechanism. Looking at Table \ref{data statistics}, Restaurants has the smallest “SQL pattern count” (i.e., lowest diversity), second largest “Questions per unique SQL query” (i.e., second highest redundancy), close to the smallest “Unique tables per query” and “SELECTs per query” (i.e., close to lowest complexity), which makes models easily learn from this dataset (Section \ref{benchmark}). For other datasets, they have higher diversity, lower redundancy, or higher complexity, which makes models harder to make predictions and the performance is generally lower than Restaurants. (2) The impact of dataset size: Smaller datasets tend to have lower performance, as shown in Table \ref{major results}, which means they are harder to learn in general due to lack of data; however, they can benefit more from our proposed FL paradigm.

%% file: related_work-v1.tex
\section{Related Work}

\noindent \textbf{Text-to-SQL.} Text-to-SQL problem which translates natural language questions to SQL queries has been studied for many years. There have been several single-database text-to-SQL datasets such as Geoquery \citep{data-atis-geography-scholar} and ATIS \citep{data-atis-geography-scholar}\nop{ Advising \cite{data-sql-advising}, Scholar \citep{data-atis-geography-scholar}, IMDB \citep{data-sql-imdb-yelp},Yelp \citep{data-sql-imdb-yelp}, Restaurants \citealp{data-restaurants-logic, data-restaurants-original,data-restaurants} and Academic \citep{data-academic} {\cl{I assume we have already listed all these refs before, if so, please consider to delete here to save space}}}, which map from natural language questions to SQL queries on a single database. \citealp{finegan-dollak-etal-2018-improving} curate eight datasets to unify their SQL format. These datasets cover a variety of domains and have different characteristics of the tables and SQL, which provide us a foundation to study the heterogeneous FL for the text-to-SQL problem.

One line of work designs special models for the text-to-SQL task such as designing a relation-aware self-attention mechanism for the Transformer model to better encode the relation of the column mappings \citep{wang-etal-2020-rat} or adding constraints to the decoder \citep{Scholak2021:PICARD} to generate valid SQL queries, while another line of work tries to directly finetune a pre-trained language model such as T5 \cite{xie-etal-2022-unifiedskg, raffel2020exploring, rajkumar2022evaluating}. As directly finetuning T5 has shown great performance and allows us to use a unified model architecture for all clients and the server\nop{it is\nop{\cl{if we want to keep this sentence, we need to use 'it is'}} also easy to unify all the clients\nop{\cl{this sentence could also be deleted}}}, we choose T5-base as the backbone model in our work.

\noindent \textbf{Heterogeneity in Federated Learning.} Heterogeneity is one of the major challenges in federated learning. Existing work \cite{pmlr-v54-mcmahan17a, reddi2020adaptive, li2020federated, li2021fedbn, shoham2019overcoming, t2020personalized,li2022federated} shows that heterogeneity can cause performance degradation. Several methods have been proposed to address this issue. For instance, FedOPT \cite{reddi2020adaptive} uses powerful adaptive optimization methods for both the server and clients, while FedProx \cite{li2020federated} (and pFedMe \cite{t2020personalized}) regularizes the local training procedure. However, based on our observations in Section \ref{main results section}, our mechanism significantly outperforms these methods. Other work that aims to address the heterogeneity issue in FL includes FedNova \cite{wang2020tackling} and \citealp{li2019fair}. Specifically, FedNova \cite{wang2020tackling} uses the local training update steps to normalize the server aggregation, and \citealp{li2019fair} proposes to optimize the power-scaled training objective. Compared to FedNova, we use a more direct indicator, training loss reduction, to adjust the weight for each client during aggregation. Different from \citealp{li2019fair}, our proposed  simple yet effective mechanism does not require modification of the local client optimization step or additional tuning of any related hyperparameter. 


%% file: conclusion.tex
\section{Conclusions}

To the best of our knowledge, we are the first to study federated learning for semantic parsing. Specifically, we propose a realistic benchmark by re-purposing eight single-domain text-to-SQL datasets.\nop{which evaluates individual clients' performance as well as their averaged performance} Moreover, we propose a novel loss reduction adjusted re-weighting mechanism (\texttt{Lorar}) that is applicable to widely adopted FL algorithms. By applying \texttt{Lorar} to FedAvg, FedOPT and FedProx, we observe their performance can be improved substantially on average, and clients with smaller datasets enjoy larger performance gains.